\documentclass[english]{article}
\usepackage{geometry}

\usepackage[T1]{fontenc}
\usepackage[latin9]{inputenc}
\usepackage{bm}
\usepackage{amsmath,mathtools}
\usepackage{amssymb}
\usepackage[unicode=true,
 bookmarks=false,
 breaklinks=false,pdfborder={0 0 1},colorlinks=false]
 {hyperref}
\hypersetup{
 colorlinks,citecolor=blue,filecolor=blue,linkcolor=blue,urlcolor=blue}

\makeatletter
\usepackage{amsthm}
\usepackage{cite}  
\usepackage{comment}
\usepackage{natbib}
\usepackage{booktabs}

\usepackage{graphicx}

\usepackage[linesnumbered,ruled,vlined]{algorithm2e}

\SetCommentSty{mycommfont}
\usepackage{algorithmic}

\usepackage{float}
\usepackage{multirow}
 
\usepackage{dsfont}
\usepackage{tcolorbox}

\allowdisplaybreaks

    \DeclareMathOperator{\ind}{\mathds{1}}  

\newtheorem*{theorem*}{Theorem}

\newcommand{\mymid}{\,|\,}

\definecolor{yanxi}{RGB}{0,200,100}

\usepackage{lipsum}

\title{Optimal Convergence Analysis of DDPM for General Distributions}

\author{%
Yuchen Jiao \thanks{Department of Statistics and Data Science, Chinese University of Hong Kong.}
\and
Yuchen Zhou \thanks{Department of Statistics, University of Illinois Urbana-Champaign.}
\and
Gen Li \footnotemark[1] 
}

\date{October 2025;~~ Revised: December 2025}

\makeatother

\begin{document}

\theoremstyle{plain} \newtheorem{lemma}{\textbf{Lemma}}\newtheorem{proposition}{\textbf{Proposition}}\newtheorem{theorem}{\textbf{Theorem}}

\theoremstyle{assumption}\newtheorem{assumption}{\textbf{Assumption}}
\theoremstyle{remark}\newtheorem{remark}{\textbf{Remark}}
\theoremstyle{definition}\newtheorem{definition}{\textbf{Definition}}
\newtheorem{example}{\textbf{Example}}

\maketitle 

\begin{abstract}
Score-based diffusion models have achieved remarkable empirical success in 
generating high-quality samples from target data distributions.
Among them, the Denoising Diffusion Probabilistic Model (DDPM) is one of the 
most widely used samplers, generating samples via estimated score functions.
Despite its empirical success, a tight theoretical understanding of 
DDPM --- especially its convergence properties --- remains limited.

In this paper, we provide a refined convergence analysis for the DDPM sampler 
and establish near-optimal convergence rates under general distributional assumptions. 
Specifically, we introduce a relaxed smoothness condition parameterized 
by a constant $L$, which is small for many practical distributions (e.g., Gaussian mixture models).
We prove that, to approximate a target distribution on $\mathbb{R}^d$ to accuracy $\varepsilon$ in total variation distance and $\varepsilon^2$ in KL-divergence, the DDPM sampler with accurate score estimates requires at most
$$T = \widetilde{O}\left(\frac{\sqrt{d}\min\{\sqrt{d},L\}}{\varepsilon}\right)$$
iterations, where $\widetilde{O}$ hides polylogarithmic factors in $d$ and $1/\varepsilon$.
This result substantially improves upon the 
best-known $\widetilde{O}(d/\varepsilon)$ iteration complexity when $L < \sqrt{d}$. 
By establishing a matching lower bound, we show that our convergence analysis is tight for a wide array of target distributions.
Moreover, it reveals that DDPM and DDIM share the same dependence on $d$, 
raising an interesting question of why DDIM often appears empirically faster. 
\end{abstract}

\section{Introduction}
The past few years have witnessed the emergence of diffusion models as a 
leading generative paradigm, achieving top performance across a wide range 
of applications, including images \citep{Rombach2021HighResolution,Saharia2022Photorealistic,Song2019Generative,Croitoru2023Diffusion,lugmayr2022repaint,nichol2021glide}, 
audio \citep{villegas2022phenaki,liu2023audioldm}, and text \citep{Ramesh2022Hierarchical,li2022diffusion}. 
In short, they consist of two components: 
\begin{itemize}
	\item[(1)] a forward process $$X_0 \sim p_{{\sf data}} \to X_1 \to \cdots \to X_T$$ that starts with $X_0$ drawn from the target data distribution and, by sequentially adding noise, yields $X_T \approx \mathcal{N}(0, I_d)$; and 
	\item[(2)] a backward process $$Y_T \sim \mathcal{N}(0, I_d) \to Y_{T-1} \to \cdots \to Y_0$$ that successively 
	transforms Gaussian noise into $Y_0$ whose distribution is close to 
	$p_{{\sf data}}$ using learned scores $s_t \approx s_t^\star = \nabla \log p_{X_t}$ \citep{Hyvarinen2005Estimation,Ho2020Denoising,Hyvrinen2007Some,Vincent2011AConnection,Song2019Generative,Pang2020Efficient}. 
\end{itemize} 
Along the trajectory, the backward marginals track the forward ones: 
for $0 \leq t \leq T$, $Y_t \stackrel{\rm{d}}{\approx} X_t$. 
To this end, there are two mainstream approaches for constructing the 
reverse process: SDE-based samplers and ODE-based samplers, 
prototyped by the Denoising Diffusion Probabilistic Model (DDPM) \citep{Ho2020Denoising}
and the Denoising Diffusion Implicit Model (DDIM) \citep{song2021denoising}, respectively. 
They differ in that the former adds an independent Gaussian perturbation when updating $Y_{t-1}$ from $Y_t$, whereas the latter is deterministic. 

Motivated by their remarkable empirical success, the convergence behavior 
of both sampler classes, particularly DDPM and DDIM, has attracted increasing theoretical attention 
\citep{Bortoli2022Convergence,Gao2023Wasserstein,Lee2022Convergence1,
Lee2022Convergence2,Chen2022Sampling,Benton2023Nearly,Chen2023Improved,
Li2023Towards,li2024sharp,Gupta2024Faster,Chen2023The,li2024accelerating,
li2024adapting,li2024d,huang2024denoising,li2024provable,li2024improved,
huang2024convergence,huang2024reverse,liang2025low,holzmuller2023convergence,
li2024unified,gao2024convergence,xu2024provably,li2025faster,li2025convergence, li2025dimension,jiao2025towards,liprovable}. 
A common objective is to control the total variation (TV) distance between the distribution generated by the sampler and the target. For a variant of the DDIM sampler, \cite{Chen2023The} established an $\widetilde{O}(\widetilde{L}^2\sqrt{d}/T)$\footnote{Throughout the paper, we say $f(L, d, T) = O(g(L, d, T))$ or $f(L, d, T) \lesssim g(L, d, T)$ if $|f(L, d, T)| \leq Cg(L, d, T)$ holds for some universal constant $C > 0$; and $f(L, d, T) \asymp g(L, d, T)$ if both $f(L, d, T) = O(g(L, d, T))$ and $g(L, d, T) = O(f(L, d, T))$ hold. The notation $\widetilde{O}(L, d, T)$ means $O(L, d, T)$ up to polylogarithmic factors.} TV rate 
 under a bounded second moment assumption on the target distribution and $\widetilde{L}$-Lipschitz conditions on both the true score $s_t^\star(\cdot)$ and its estimator $s_t(\cdot)$. For DDPM, the best-known rate is $\widetilde{O}(d/T)$ \citep{li2024d}, which is much worse than the DDIM bound when $\widetilde{L} \ll d^{1/4}$. This natually leads to the following questions:

\begin{itemize}
	\item[] \emph{Under comparable or even weaker assumptions, can DDPM achieve an $\widetilde{O}(\sqrt{d}/T)$ rate in TV distance, or $\widetilde{O}(d/T^2)$ rate in Kullbeck-Leibler (KL) divergence, for a broad class of target distributions?} 
\end{itemize}

\paragraph{Main contributions.}
We summarize our main results in the following informal theorem; a formal version is deferred to Section~\ref{sec:theory}.
\begin{theorem*}[Informal]
	Suppose that the score functions associated with the forward process satisfies a relaxed Lipschitz condition with parameter $L$ (see Definition~\ref{def:score-lipschitz}), a much smaller quantity than the global Lipschitz constant commonly assumed in prior work. Then, with access to sufficiently accurate score estimates, the convergence rate of the DDPM sampler in total variation (resp.~in  KL divergence) is 
	$$\widetilde{O}\Big(\frac{\sqrt{d}\min\{L, \sqrt{d}\}}{T}\Big)~\quad~(\text{resp.}~\widetilde{O}\Big(\frac{d\min\{L^2, d\}}{T^2}\Big)).$$ 	
\end{theorem*}
We next discuss the implications of this result.

\begin{enumerate}
	\item {\bf Optimal $\sqrt{d}$ dependence for common distributions}. 
	We show that our relaxed Lipschitz condition is satisfied by many common target distributions with $L \le \mathrm{poly}(\log(dT))$, and consequently the convergence rate in total variation distance (resp.~KL divergence) becomes $$\widetilde{O}\Big(\frac{\sqrt{d}}{T}\Big)~\quad~(\text{resp.}~\widetilde{O}\Big(\frac{d}{T^2}\Big)).$$
	Our results significantly reduce the dependence on $d$ in DDPM convergence analyses. Given that the ambient dimension $d$ often scales as $10^4 - 10^6$ in common image/video tasks, this improvement is substantial. Compared with the $\widetilde{O}(\widetilde{L}^2\sqrt{d}/T)$ rate for DDIM, our result shares the same $d$ dependence while improving the dependence on the Lipschitz parameter $L$. We further established matching lower bounds, implying that our TV and KL rates are tight (up to logarithmic factors). 
	\item {\bf $\widetilde{O}(d^2/T^2)$ KL rate under minimal assumptions}.  Even without the non-uniform Lipschitz assumption (i.e., allowing $L=\infty$), our analysis yields upper bounds of order $\widetilde{O}(d/T)$ in TV distance and $\widetilde{O}(d^2/T^2)$ in KL divergence under the minimal assumption on the target distribution (Assumption~\ref{assu:distribution}). This matches the state-of-the-art results in \cite{li2024d} and \citet{jain2025sharp} for the TV- and KL-rate analysis, respectively.
\end{enumerate}

\paragraph{Paper organization.} The rest of the paper is organized as follows. In Section~\ref{sec:preliminary}, we formally introduce diffusion models and the DDPM sampler. The main results are presented in Section~\ref{sec:main_theory}. In Section~\ref{sec:analysis}, we outline the proof of of Theorem~\ref{thm:main}. The proofs of all technical lemmas are deferred to the appendix.

\section{Preliminary}\label{sec:preliminary}
In this section, we introduce preliminaries on diffusion models and the DDPM sampler.

\paragraph{The forward process.}
Diffusion models comprise a forward (noising) process and the reverse (denoising) process.
The forward process is Markovian:
starting from
$X_0\sim p_{\mathsf{data}}$ on $\mathbb{R}^d$, 
\begin{align}\label{eq:forward_process}
	X_{t} = \sqrt{\alpha_t}X_{t-1} + \sqrt{1-\alpha_t} Z_t,\quad  t=1,\cdots,T,
\end{align}
where $Z_t \stackrel{\text{i.i.d.}}{\sim}\mathcal{N}(0,I_d)$ are independent Gaussian random vectors, and the $\alpha_t$'s are the learning rates.
Then it is straightforward to show that
\begin{align}\label{eq:forward}
	X_{t} = \sqrt{\overline{\alpha}_t}X_0 + \sqrt{1-\overline{\alpha}_t} ~\overline{Z}_t,\quad \mathrm{where}\quad 
	\overline{\alpha}_t = \prod_{k=1}^t\alpha_k~\text{and}~\overline{Z}_t\sim\mathcal{N}(0,I_d).
\end{align}

In the continuous-time limit, the forward process admits the following widely studied SDE:
\begin{align}\label{eq:forward-ODE}
	\mathrm{d} \overline{X}_\tau = -\frac{1}{2(1-\tau)} \overline{X}_\tau \mathrm{d}\tau + \frac{1}{\sqrt{1-\tau}} \mathrm{d} B_\tau,~\quad~\tau\in(0,1),~\quad~\overline{X}_0 \sim p_{\mathsf{data}},
\end{align}
where $B_\tau$ is the standard Brownian motion.
In fact, it has been shown that the distribution of $\overline{X}_\tau$ is given by
\begin{align}\label{eq:forward-distribution-Xtau}
\overline{X}_\tau \overset{\text{d}}{=} \sqrt{1-\tau}~\overline{X}_0 + \sqrt{\tau} W,\quad W\sim\mathcal{N}(0,I_d).
\end{align}
Putting \eqref{eq:forward} and \eqref{eq:forward-distribution-Xtau} together,
we establish the following connection between the forward processes in discrete time and continuous time:
\begin{align}\label{eq:tau_t}
	\overline{X}_{\tau_{T-t+1}} \overset{\text{d}} {=} X_t,\quad \mathrm{where}\quad \tau_{T-t+1} = 1-\overline{\alpha}_t. 
\end{align}
Throughout this paper, we denote by $p_{X_t}$ and $p_{\overline{X}_\tau}$ the probablity density function of $X_t$ and $\overline{X}_\tau$, respectively.

\paragraph{Score functions.}
A key ingredient 
in the sampling process is the (Stein) score function, i.e., the gradient of the log marginal density of the forward process~\eqref{eq:forward_process}:
\begin{align}\label{eq:score_discrete}
	s_t^{\star}(x) \coloneqq \nabla\log p_{X_t}(x) = -\frac{1}{1-\overline{\alpha}_t}\int (x -\sqrt{\overline{\alpha}_t}x_0)p_{X_0|X_t}(x_0| x)\mathrm{d}x_0,
\end{align}
where the last equality follows from Tweedie's formula \citep{efron2011tweedie}.
For notational convenience, we also introduce the score for the continuous-time process \eqref{eq:forward-ODE}:
\begin{align}\label{eq:score_continuous}
	\overline{s}_\tau^{\star}(x) \coloneqq \nabla\log p_{\overline{X}_\tau}(x) = -\frac{1}{\tau}\int (x -\sqrt{1-\tau}x_0)p_{\overline{X}_0|\overline{X}_{\tau}}(x_0| x)\mathrm{d}x_0.
\end{align}
Recalling that $\overline{X}_{\tau_{T-t+1}}\overset{\rm d}{=}X_t$ (cf.~\eqref{eq:tau_t}), we immediately have $\overline{s}_{\tau_{T-t+1}}^{\star}(x) = s_{t}^{\star}(x)$.

We usually do not have access to the exact score functions $s_t^\star(\cdot)$. Here, we assume that some estimates for the score functions, $\{s_t(\cdot)\}_{1 \leq t \leq T}$, are available.

\paragraph{DDPM Sampler and learning rate schedule.}
The Denoising Diffusion Probabilistic Model (DDPM) constructs a reverse process for \eqref{eq:forward}, with the goal of generating samples whose distribution is close to $p_{\mathsf{data}}$. 
More specifically, each $Y_{t-1}$ is a function of $Y_t$ plus independent Gaussian noise $Z_t \sim \mathcal{N}(0, I_d)$:
\begin{align}\label{eq:DDPM}
	Y_{t-1} = \frac{1}{\sqrt{\alpha_t}}\left(Y_t + \eta_t s_t(Y_t) + \sigma_tZ_t\right), \quad t=T,\cdots,2,\quad Y_T\sim \mathcal{N}(0,I_d)
\end{align}
where $\eta_t$ and $\sigma_t$ are parameters that play pivotal roles for achieving satisfactory performance. 
Following \citet{li2024sharp}, we choose 
\begin{align}\label{eq:def-eta-sigma}
	\eta_t = 1-\alpha_t,
	\qquad \sigma_t^2 = 1-\alpha_t.
\end{align}
We assume that the learning rates satisfy the following conditions: (i) $\beta_t := 1 - \alpha_t$ is small for every $1 <t<T$; 
and (ii) $\overline{\alpha}_T = \prod_{t=1}^T\alpha_t$ is vanishingly small, ensuring that the distribution of $X_T$, 
is exceedingly close to $\mathcal{N}(0, I_d)$. 
More specifically, we assume that the learning rates $\{\alpha_{t}\}_{1 \leq t \leq T}$ satisfy
\begin{align}\label{eq:learning-rate}
	\overline{\alpha}_T\le \frac{1}{T^{c_0}},\qquad 1-\alpha_1 \le \frac{1}{T^{c_0}},\qquad \overline{\alpha}_{t-1} - \overline{\alpha}_t \le \frac{c\log T}{T}\overline{\alpha}_t(1-\overline{\alpha}_t).
\end{align}
An example of learning rate schedule that obeys \eqref{eq:learning-rate} is:
\begin{align}\label{eq:learning-rate-example}
	\beta_1 = \frac{1}{T^{c_0}},\qquad \beta_{t+1} = \frac{c_1\log T}{T}\min\left\{\beta_1\left(1+\frac{c_1\log T}{T}\right)^t,1\right\},\qquad t = 1,\cdots, T-1.
\end{align}

\paragraph{Notation}  For two probability measures $P$ and $Q$,  the total-variation distance between them is defined as ${\sf TV}(P,Q) \coloneqq \frac{1}{2}\int|{\rm d}P - {\rm d}Q|$. If $P$ is absolutely continuous with respect to $Q$, the  Kullbeck-Leibler (KL) divergence of $P$ from $Q$ is ${\sf KL}(P,Q) \coloneqq \int\log(\frac{{\rm d}P}{{\rm d}Q}){\rm d}P$.
For any random vector $X$, we let $p_X$ denote its probability density function. For any matrix $A$, we denote by $\|A\|_{\mathsf{op}}$ its spectral norm. 


\section{Main theory}\label{sec:main_theory}
In this section, we develop an optimal convergence rate theory for the DDPM sampler, 
showing that it has the same $O(\sqrt{d})$ dependence as DDIM.
Before proceeding, we first introduce the assumptions used in our analysis.

\subsection{Assumptions}

The first assumption allows the second-order moment of the target distribution to scale at most polynomially in $T$, which encompasses a wide array of applications.
\begin{assumption}\label{assu:distribution}
	The target distribution $p_{\mathsf{data}}$ has a bounded second-order moment:
	\begin{align}\label{assump:bounded_second_moment}
		\mathbb{E}_{X_0\sim p_{\mathsf{data}}}[\|X_0\|_2^2] \leq T^{c_R},
	\end{align}
	where $c_R > 0$ is an arbitrarily large constant.
\end{assumption}

Our analysis makes use of a relaxed smoothness condition. Specifically, we introduce a non-uniform Lipschitz constant for the normalized score functions $\tau \overline{s}_{\tau}^{\star}$ as follows:
\begin{definition}[Non-uniform Lipschitz property]\label{def:score-lipschitz}
	Let $L \geq 1$ denote the smallest quantity such that, for every $\tau\in(0,1)$,
	\begin{align*}
		\mathbb{P}\left(\tau\|\nabla \overline{s}_\tau^{\star}(\overline{X}_\tau)\|_{\mathsf{op}}\le L\right) \ge 1-\frac{1}{d^4}.
	\end{align*}
\end{definition}
Compared to the global smoothness condition $\|\nabla \overline{s}_\tau^{\star}(x)\|_{\mathsf{op}}\le \widetilde{L}$ for all $x\in\mathbb{R}^d$, 
which is widely used in diffusion-model analyses,
our assumption is milder and applies to a broader range of data distributions: (1) it only requires a high-probability bound, and (2) it bounds the scaled quantity $\tau\|\nabla \overline{s}_\tau^{\star}(\overline{X}_\tau)\|_{\mathsf{op}}$, thereby permitting $\|\nabla \overline{s}_\tau^{\star}(\overline{X}_\tau)\|_{\mathsf{op}}$ to be much larger when $\tau$ is small (when the distribution of $X_{\tau}$ is closer to the target $p_{\rm data}$). The following examples show that the non-uniform Lipschitz property holds for many common distributions with $L \lesssim \log(dT)$, whereas the global smoothness condition may fail (i.e., $\widetilde{L} = \infty$).
\begin{example}\label{example:gauss}
	Suppose that $X_0 \sim \mathcal{N}(\mu,\Sigma)$ for any $\mu\in \mathbb{R}^d$ and covariance matrix $\Sigma$. 
	Then the non-uniform Lpischitz $L = 1$.
\end{example}

\begin{example}\label{example:GMM}
	Suppose that $X_0$ follows a $d$-dimensional Gaussian Mixture Model $$\sum_{h=1}^H \pi_h \mathcal{N}(\mu_h,\sigma_h^2I_d),~\quad~\pi_h \geq 0,~\quad~\sum_{h=1}^{H}\pi_h = 1.$$
	Then the non-uniform Lipischitz constant obeys $L\lesssim \log(d)\log(H)$.
\end{example}

\begin{example}\label{example:independent}
	Suppose that the entries of $X_0$ are independent and satisfy $\mathbb{E}[|X_{0,i}|]\le d^{c_R}$,
	where $X_{0,i}$ is the $i$-th coordinate and $c_R$ is an arbitrarily large constant.
	Then the non-uniform Lipschitz constant satisfies $L\lesssim \log d$.
\end{example}
In fact, we conjecture that the non-uniform Lipschitz condition holds with $L \le \mathrm{poly}(\log(dT))$ for any absolutely continuous target distribution. We leave further investigation on this for future work. 

In addition, the following assumption captures the quality of the score estimates.
\begin{assumption}\label{assu:score-error}
	We assume access to an estimate $s_t(\cdot)$ for each $s^{\star}_t(\cdot)$, with the averaged $\ell_2$ score estimation error
	\begin{align} \label{eq:score-error}
		\varepsilon_{\mathsf{score}}^2 &\coloneqq \frac{1}{T}\sum_{t = 1}^{T-1} (1-\overline{\alpha}_t)\mathbb{E}_{x_{t}\sim p_{X_t}}\big[\|s_{t}(x_t) - s_{t}^{\star}(x_t)\|_2^2\big],
	\end{align}
	where $\overline{\alpha}_t$ satisfies \eqref{eq:learning-rate}.
\end{assumption}
Note that $\varepsilon_{\mathsf{score}}^2$ is no larger than the commonly used unweighted error  \citep{Gupta2024Faster,huang2024convergence,huang2024reverse,Lee2022Convergence1,li2024provable,li2024accelerating}
\begin{align}\label{eq:score-error-unweighted}
	\widetilde{\varepsilon}_{\mathsf{score}}^2 := \frac{1}{T}\sum_{t = 1}^{T-1}\mathbb{E}_{x_{t}\sim p_{X_t}}\big[\|s_{t}(x_t) - s_{t}^{\star}(x_t)\|_2^2\big]
\end{align}
since $0<1-\overline{\alpha}_t\le 1$ for all $t$. In particular, under the learning-rate schedule \eqref{eq:learning-rate}, the weights $1-\overline{\alpha}_t$ can be very small for early steps, so Assumption~\ref{assu:score-error} allows larger per-step errors $\mathbb{E}\|s_t(X_t)-s_t^{\star}(X_t)\|_2^{2}$ when $t$ is small.

\subsection{Theory}\label{sec:theory}

The following theorem establishes a sharp convergence rate for the classical DDPM sampler.
\begin{theorem}\label{thm:main}
	Suppose that Assumptions \ref{assu:distribution} and \ref{assu:score-error} hold.  
	Then the DDPM sampler \eqref{eq:DDPM} with the learning rate schedule \eqref{eq:learning-rate} satisfies
	\begin{align}\label{eq:TV-main}
		\mathsf{TV}\left( p_{X_1},p_{Y_1}\right)\le \sqrt{\frac12\mathsf{KL}\left( p_{X_1}\Vert p_{Y_1}\right)} \le \frac{Cd^{1/2}\min\{d^{1/2}, L\}\log^{2} T}{T} + C\varepsilon_{\mathsf{score}}\log^{1/2} T,
	\end{align}
	for some constant $C > 0$ large enough, where $L$ is defined in Definition \ref{def:score-lipschitz}.
\end{theorem}

Assume that the perfect scores are available, i.e., $\varepsilon_{\mathsf{score}} =0$.
Then the convergence rate in total variation distance (resp.~KL divergence) becomes
$$
\widetilde{O}\left(\frac{d^{1/2}\min\{d^{1/2},L\}}{T}\right)~\quad~(\text{resp.}~\widetilde{O}\Big(\frac{d\min\{d,L^2\}}{T^2}\Big)).
$$
When $L < \sqrt{d}$, our results substantially improve upon the state-of-the-art $\widetilde{O}(d/T)$ TV rate \citep{li2024d} and $\widetilde{O}(d^2/T^2)$ KL rate \citep{,jain2025sharp}. Moreover, our analysis demonstrates that DDPM achieves an $O(\sqrt{d})$  dependence on dimension, matching (a variation of) DDIM,
while enjoying a better dependence on the Lipschitz constant. 
Specifically, our bound is linear in $L$, versus the quadratic $O(\widetilde{L}^2)$ dependence established for DDIM \citep{Chen2023The}. (we recall that $\widetilde{L}$ is the global smoothness constant, which is larger than our non-uniform $L$). For the score-error term, whereas prior work incurs $O(\widetilde{\varepsilon}_{\rm score}\log^{1/2}T)$ (with $\widetilde{\varepsilon}_{\rm score}$ defined in \eqref{eq:score-error-unweighted}), our TV bound depends on $\varepsilon_{\rm score}\log^{1/2}T$, which is tighter since $\widetilde{\varepsilon}_{\rm score}\geq \varepsilon_{\rm score}$. 

In addition, we can prove the following lower bound, which confirms the tightness of Theorem~\ref{thm:main} when $L \leq \text{poly}(\log(dT))$.
The proof is postponed to Appendix \ref{sec:proof-thm-lower-bound}.
\begin{theorem}\label{thm:lower_bound}
	Assume that the learning rates $\{\alpha_t\}_{t=1}^T$ satisfy \eqref{eq:learning-rate} and $\eta_t=1-\alpha_t$ for all $t$, and assume  $\{\sigma_t^2\}_{t=1}^T$ satisfy one of the following: (i) $\sigma_t^2=1-\alpha_t$, (ii) $\sigma_t^2=(1-\alpha_t)\alpha_t$, and (iii) $\sigma_t^2=\frac{(\alpha_t-\overline{\alpha}_t)(1-{\alpha}_{t})}{1-\overline{\alpha}_t}$. 
	If $p_{\text{data}} = \mathcal{N}(0,\lambda I_d)$ with some constant $\lambda \geq 2$,
	the output of sampler \eqref{eq:DDPM} with the oracle scores $s_t(\cdot) = s_t^\star(\cdot)$ obeys:
	\begin{align*}
	\mathsf{KL}(p_{X_1}||p_{Y_1}) \ge \frac{c_{\mathsf{low}}d}{T^2}
	\end{align*}
	for some universal constant $c_{\mathsf{low}} > 0$.
\end{theorem}
\begin{remark}
	Theorem~\ref{thm:lower_bound} focuses on the learning-rate schedules most widely used in the literature \citep{Ho2020Denoising,li2024d,Li2023Towards,huang2024denoising}.
	Extending the analysis to arbitrary schedules is left for future investigation.
\end{remark}



\section{Analysis}
\label{sec:analysis}

The proof is divided into three steps.

\paragraph{Step 1. constructing an auxiliary reverse process.}
To begin with, we introduce an operator $$\Phi_{\tau_1\to\tau_2}(x)\coloneqq x_{\tau_2}|_{x_{\tau_1} = x},$$ which is defined via the following ODE:
\begin{align}\label{eq:ODE}
	\mathrm{d}\frac{x_\tau}{\sqrt{1-\tau}} = -\frac{\overline{s}_\tau^{\star}(x_\tau)}{2(1-\tau)^{3/2}}\mathrm{d} \tau.
\end{align}
By virtue of \citet[Eqn.~(20)]{li2024accelerating}, 
we know that
\begin{align}\label{eq:ODE-d}
	\Phi_{\tau_1\to\tau_2}(\overline{X}_{\tau_1}) \overset{\rm{d}}{=} \overline{X}_{\tau_2}.
\end{align}
Motivated by this fact, we construct the following auxiliary reverse process 	
\begin{align}\label{eq:def-hatX}
	\widehat{X}_T &\sim p_{X_T},\notag\\
	\widehat{X}_{t-1} 
	&:= \frac{\sqrt{\overline{\alpha}_{t-1}}}{\sqrt{1-\widehat{\tau}_{T-t+2}}}\Phi_{\tau_{T-t+1}\to\widehat{\tau}_{T-t+2}}(\widehat{X}_t) + \frac{\sigma_t}{\sqrt{\alpha_t}}Z_t\notag\\
	&=\frac{1}{\sqrt{\alpha_t}}\widehat{X}_t + \int_{\widehat{\tau}_{T-t+2}}^{\tau_{T-t+1}}\frac{\sqrt{\overline{\alpha}_{t-1}}\overline{s}_\tau^{\star}(x_\tau)}{2(1-\tau)^{\frac32}}\mathrm{d}\tau + \frac{\sigma_t}{\sqrt{\alpha_t}} Z_t,\quad t=T,\cdots,2.
\end{align}
Here, $Z_t \stackrel{\rm{i.i.d.}}{\sim}\mathcal{N}(0,I_d)$ are independent Gaussian random vectors, $\tau_{T - t + 1} = 1 - \overline{\alpha}_t$ is defined in \eqref{eq:tau_t}, and the $x_\tau$'s are defined via the ODE \eqref{eq:ODE} with $x_{\tau_{T-t+1}} = \widehat{X}_t$, and 
\begin{align}\label{eq:def-hattau}
	\widehat{\tau}_{T-t+2} = 1 - \widehat{\alpha}_{t-1},\quad \mathrm{where}\quad \widehat{\alpha}_{t-1}\coloneqq \frac{\overline{\alpha}_{t}}{2\alpha_t-1}.
\end{align}
 The following lemma shows that $\widehat{X}_t$ and the forward process $X_t$ share the same marginal distribution, with the proof deferred to Appendix \ref{sec:proof-lem-ODE}.

\begin{lemma}\label{lem:ODE}
	For all $1\le t\le T$, we have
	\begin{align*}
		\widehat{X}_{t} \overset{\rm{d}}{=} X_t, 
	\end{align*}
	where $\widehat{X}_{t}$ and $X_t$ are defined in \eqref{eq:def-hattau} and \eqref{eq:forward}, respectively.
\end{lemma}

In view of Pinsker's inequality and Lemma \ref{lem:ODE}, we have
\begin{align}
	\mathsf{TV}^2(p_{{Y}_{1}},p_{{X}_{1}})
	&\le \frac12\mathsf{KL}\left(p_{{X}_{1}}\Vert p_{{Y}_{1}}\right) = \frac12\mathsf{KL}\left(p_{\widehat{X}_{1}}\Vert p_{{Y}_{1}}\right)\nonumber\\
	&\overset{\text{}}{\le} \frac12\mathsf{KL}\left(p_{\widehat{X}_{T}}\Vert p_{{Y}_{T}}\right) + \frac{1}{2}\sum_{t=2}^{T}\mathbb{E}_{x_t\sim\widehat{X}_t}\mathsf{KL}(p_{\widehat{X}_{t-1}\mymid\widehat{X}_t}(\cdot\mymid x_t)\Vert p_{{Y}_{t-1}\mymid{Y}_t}(\cdot\mymid x_t)).\label{eq:proof-decomp-error}
\end{align}

Here, the last inequality makes use of the chain rule of KL divergence. 

\paragraph{Step 2. controlling discretization and estimation error.}

This step focuses on bounding the term $$
\sum_{t=2}^{T}\mathbb{E}_{x_t\sim\widehat{X}_t}\mathsf{KL}(p_{\widehat{X}_{t-1}\mymid\widehat{X}_t}(\cdot\mymid x_t)\Vert p_{{Y}_{t-1}\mymid{Y}_t}(\cdot\mymid x_t))
$$ 
appearing on the right-hand side of \eqref{eq:proof-decomp-error}. Eqn.~\eqref{eq:def-hattau} tells us that the conditional density of $\widehat{X}_{t-1}$ given $\widehat{X}_{t} = x_t$ is
\begin{align}\label{eq:trans-prob-hatX}
	p_{\widehat{X}_{t-1}\mymid \widehat{X}_{t}}(x\mymid x_t) = \phi\bigg(x~\bigg |~ \frac{\sqrt{\overline{\alpha}_{t-1}}}{\sqrt{1-\widehat{\tau}_{T-t+2}}}\Phi_{\tau_{T-t+1}\to\widehat{\tau}_{T-t+2}}(x_t), \frac{\sigma_t^2}{\alpha_t}\bigg),
\end{align}
where $\phi(x|\mu,\sigma^2)$ denotes the probability density function of the $d$-dimensional Gaussian distribution with mean vector $\mu$ and covariance matrix $\sigma^2I_d$.
Moreover, from \eqref{eq:DDPM}, the conditional density of ${Y}_{t-1}$ given ${Y}_{t} = y_t$, for $t=0,\cdots,T-1$, is
\begin{align}\label{eq:trans-prob-Y}
	P_{Y_{t-1}|Y_t}(y|{y}_t) = \phi\left(y~\bigg |~\frac{1}{\sqrt{\alpha_t}}(y_t + \eta_t s_t(y_t)),\frac{\sigma_t^2}{\alpha_t}\right).
\end{align}
In other words, both $\widehat{X}_{t-1}$ and $Y_{t-1}$ are conditionally Gaussian given $\widehat{X}_t = Y_t = x_t$, with the same variance $\sigma_t^2/\alpha_t$ but different mean vectors.
As a consequence, the KL divergence between $p_{\widehat{X}_{t-1}\mymid\widehat{X}_t}(\cdot\mymid x_t)$ and $p_{{Y}_{t-1}\mymid{Y}_t}(\cdot\mymid x_t)$ can be calculated as
\begin{align}\label{eq:cal-KL}
	\mathsf{KL}(p_{\widehat{X}_{t-1}\mymid\widehat{X}_t}(\cdot\mymid x_t)\Vert p_{{Y}_{t-1}\mymid{Y}_t}(\cdot\mymid x_t)) = \frac{\alpha_t}{2\sigma_t^2}\|\mu_{\widehat{X}_{t-1}|\widehat{X}_t}(x_t) - \mu_{Y_{t-1}|Y_t}(x_t)\|_2^2,
\end{align}
where the conditional mean vectors are
\begin{subequations}\label{eq:def-mu-hatX-Y}
	\begin{align}
		\mu_{\widehat{X}_{t-1}|\widehat{X}_t}(x_t) &= \frac{\sqrt{\overline{\alpha}_{t-1}}}{\sqrt{1-\widehat{\tau}_{T-t+2}}}\Phi_{\tau_{T-t+1}\to\widehat{\tau}_{T-t+2}}(x_t),\\
		\mu_{Y_{t-1}|Y_t}(x_t) &=\frac{1}{\sqrt{\alpha_t}}(x_t + \eta_t s_t(x_t)) = \frac{1}{\sqrt{\alpha_t}}(x_t + \eta_t s_t^{\star}(x_t)) + \frac{\eta_t}{\sqrt{\alpha_t}}(s_t(x_t) - s_t^{\star}(x_t)).
	\end{align}
\end{subequations}
Substituting \eqref{eq:def-mu-hatX-Y} into \eqref{eq:cal-KL} and applying the inequality $\|a+b\|_2^2\le 2\|a\|_2^2 + 2\|b\|_2^2$, we obtain
\begin{align}\label{eq:cal-KL-decomp}
	\mathsf{KL}(p_{\widehat{X}_{t-1}\mymid\widehat{X}_t}(\cdot\mymid x_t)\Vert p_{{Y}_{t-1}\mymid{Y}_t}(\cdot\mymid x_t)) &\le  \frac{\alpha_t}{\sigma_t^2}\bigg\|\mu_{\widehat{X}_{t-1}|\widehat{X}_t}(x_t) - \frac{1}{\sqrt{\alpha_t}}(x_t + \eta_t s_t^{\star}(x_t))\bigg\|_2^2 + \frac{\eta_t^2}{\sigma_t^2}\|s_t(x_t) - s_t^{\star}(x_t)\|_2^2\notag\\
	&\eqqcolon \|\xi_{t-1}(x_t)\|_2^2 + \frac{\eta_t^2}{\sigma_t^2}\|s_t(x_t) - s_t^{\star}(x_t)\|_2^2.
\end{align}
Here, the first term $\xi_{t-1}(x_t)$ represents the discretization error, and the second term represents the score estimation error.
We now proceed to analyze these two error components separately.
The following lemma gives a sharp bound on the discretization error.
\begin{lemma}\label{lem:discretization}
	Let $\Sigma_\tau(x)$ denote the conditional covariance matrix of the standard Gaussian noise $Z \sim N(0, I_d)$ conditioned on $\sqrt{1-\tau}X_0 + \sqrt{\tau}Z = x$, i.e.,
	\begin{align}\label{eq:def-Sigma}
		\Sigma_\tau(x) = \mathsf{Cov}(Z|\sqrt{1-\tau}X_0 + \sqrt{\tau}Z = x).
	\end{align}
	Then the sum of the expected discretization error $\|\xi_{t-1}(x_t)\|_2^2$ defined in \eqref{eq:cal-KL-decomp} satisfies
	\begin{align*}
		\sum_{t=2}^{T}\mathbb{E}_{x_t\sim\widehat{X}_t} [\|\xi_{t-1}(x_t)\|_2^2] \lesssim \sum_{t=2}^{T}\frac{d\overline{\alpha}_{t}\log^3 T}{T^2(1-\overline{\alpha}_{t})}\int_{\widetilde{\tau}_{T-t+2}}^{\tau_{T-t+1}}\mathbb{E}_{x_\tau\sim \overline{X}_\tau} \frac{\left\|\Sigma_\tau(x_\tau)\right\|_{\mathsf{op}}^2}{(1-\tau)^2} \mathrm{d} \tau + 
		\frac{d\min\left\{d\log T,L\right\}\log^3 T}{T^2},
	\end{align*}
    where
    \begin{align}\label{eq:def-tildetau}
	\widetilde{\tau}_{T-t+2} = 1 - \widetilde{\alpha}_{t-1},\quad \mathrm{where}\quad \widetilde{\alpha}_{t-1}\coloneqq \frac{\overline{\alpha}_{t-1}(1-\overline{\alpha}_t)}{\overline{\alpha}_{t-1}(1-\alpha_t) + \alpha_t(1-\overline{\alpha}_{t-1})}.
\end{align}
\end{lemma}
Recalling the definition of $\eta_t$ and $\sigma_t$ in \eqref{eq:def-eta-sigma}, the estimation error can be written as
\begin{align}\label{eq:proof-estimation-error-temp-1}
	\frac{\eta_t^2}{\sigma_t^2}\|s_t(x_t) - s_t^{\star}(x_t)\|_2^2 = 
    (1-\alpha_t)\|s_t(x_t) - s_t^{\star}(x_t)\|_2^2 = \frac{\overline{\alpha}_{t-1}-\overline{\alpha}_t}{\overline{\alpha}_{t-1} }\|s_t(x_t) - s_t^{\star}(x_t)\|_2^2.
\end{align}
Applying the learning rate relationship \eqref{eq:learning-rate}, it follows that
\begin{align*}
\frac{\overline{\alpha}_{t-1}-\overline{\alpha}_t}{\overline{\alpha}_{t-1} } \le \frac{c\log T}{T}(1-\overline{\alpha}_t),
\end{align*}
provided that $T\ge 2c\log T$.
Putting the previous inequality and \eqref{eq:proof-estimation-error-temp-1} together, one has
\begin{align*}
	\frac{\eta_t^2}{\sigma_t^2}\|s_t(x_t) - s_t^{\star}(x_t)\|_2^2\le \frac{c\log T}{T}(1-\overline{\alpha}_t)\|s_t(x_t) - s_t^{\star}(x_t)\|_2^2.
\end{align*}
By virtue of Lemma \ref{lem:ODE}, Assumption \ref{assu:score-error} and the previous inequality, we arrive at
\begin{align}\label{eq:bound-eps-sc}
	\sum_{t=2}^{T}\mathbb{E}_{x_t\sim\widehat{X}_t}\left[\frac{\eta_t^2}{\sigma_t^2}\|s_t(x_t) - s_t^{\star}(x_t)\|_2^2\right]
	&\le \frac{c\log T}{T}(1-\overline{\alpha}_t)\sum_{t=2}^{T}\mathbb{E}_{x_t\sim\widehat{X}_t}[\|s_t(x_t) - s_t^{\star}(x_t)\|_2^2]\notag\\
	&= \frac{c\log T}{T}(1-\overline{\alpha}_t)\sum_{t=2}^{T}\mathbb{E}_{x_t\sim{X}_t}[\|s_t(x_t) - s_t^{\star}(x_t)\|_2^2] = c\log T\varepsilon_{\mathsf{score}}^2.
\end{align}

\paragraph{Step 3. putting everything together.}
Combining \eqref{eq:proof-decomp-error}, \eqref{eq:cal-KL-decomp}, \eqref{eq:bound-eps-sc} and Lemma \ref{lem:discretization}, we obtain an upper bound on the expectation of the KL divergence:
\begin{align}\label{ineq:KL_second_last_bound}
	\mathsf{TV}^2(p_{{Y}_{1}},p_{{X}_{1}})
	&\le \frac12\mathsf{KL}\left(p_{{X}_{1}}\Vert p_{{Y}_{1}}\right)\notag\\ &\lesssim
	\mathsf{KL}(p_{\widehat{X}_{T}}\Vert p_{{Y}_{T}}) + \sum_{t=2}^T\frac{d\overline{\alpha}_{t}\log^3 T}{T^2(1-\overline{\alpha}_{t})}\int_{\widetilde{\tau}_{T-t+2}}^{\tau_{T-t+1}}\frac{\mathbb{E}_{x_\tau\sim \overline{X}_\tau} \left\|\Sigma_\tau(x_\tau)\right\|_2^2}{(1-\tau)^2}\mathrm{d} \tau + 
	\frac{d\min\left\{d\log T,L\right\}\log^3 T}{T^2} + \log T\varepsilon_{\mathsf{score}}^2.
\end{align}
To prove the desired error rate, one still needs to bound $\mathsf{KL}(p_{\widehat{X}_{T}}\Vert p_{{Y}_{T}})$ and $\sum_{t=2}^T\frac{d\overline{\alpha}_{t}\log^3 T}{T^2(1-\overline{\alpha}_{t})}\int_{\widetilde{\tau}_{T-t+2}}^{\tau_{T-t+1}}\frac{\mathbb{E}_{x_\tau\sim \overline{X}_\tau} \left\|\Sigma_\tau(x_\tau)\right\|_{\mathsf{op}}^2}{(1-\tau)^2}\mathrm{d} \tau$. The following lemma gives the required bounds.
\begin{lemma}\label{lem:endpoints}
	Suppose that the learning rates satisfy \eqref{eq:learning-rate}. Then we have
		\begin{align}
			\mathsf{KL}\left(p_{\widehat{X}_{T}}\Vert p_{{Y}_{T}}\right) &\lesssim \frac{1}{T^{10}},\qquad \mathrm{and}\qquad 
			\sum_{t=2}^T\frac{\overline{\alpha}_{t}}{(1-\overline{\alpha}_{t})}\int_{\widetilde{\tau}_{T-t+2}}^{\tau_{T-t+1}}\frac{\mathbb{E}_{x_\tau\sim \overline{X}_\tau} \left\|\Sigma_\tau(x_\tau)\right\|_{\mathsf{op}}^2}{(1-\tau)^2}\mathrm{d} \tau \lesssim\min\{L^2,d\}\log T.\label{eq:sum-int-Sigma}
		\end{align}
\end{lemma}
The proof of Lemma~\ref{lem:endpoints} is deferred to Appendix \ref{sec:proof-lem-endpoints}.  Lemma \ref{lem:endpoints} together with \eqref{ineq:KL_second_last_bound} implies
\begin{align}
	\mathsf{TV}(p_{X_1},p_{Y_1}) 
	&\le \sqrt{\frac12 \mathsf{KL}(p_{X_1}\Vert p_{Y_1})}\notag\\
	&\lesssim \frac{1}{T^5} + \frac{d^{1/2}\min\{d^{1/2},L\}\log^2T}{T} + \frac{d^{1/2}\min\{d^{1/2}\log^{1/2}T,L^{1/2}\}\log^{3/2}T}{T}+ \varepsilon_{\mathsf{score}}\log^{1/2}T\notag\\
	&\lesssim \frac{d^{1/2}\min\{d^{1/2},L\}\log^2T}{T} + \varepsilon_{\mathsf{score}}\log^{1/2}T .
\end{align}

\section{Discussion}
This paper provides an optimal convergence analysis of the DDPM sampler. Under a minimal assumption on the second moment of the target distribution and a mild non-uniform Lipschitz assumption, we show that the DDPM sampler's convergence rate scales as $O(\sqrt{d})$, which significantly improves the state-of-the-art linear dependence on $d$. Our convergence guarantees in both total variation and KL divergence are tight for a broad class of distributions. These results also challenge the prevailing view that DDPM is inherently slower than DDIM due to linear dependence on $d$: we prove that DDPM and DDIM share the same $d$-dependence, and the empirical advantage of DDIM remains an open question.

Beyond this work, there are many future directions that are worth investigating. For example, our theory focuses on the original DDPM sampler; it would be interesting to see whether our analysis framework can sharpen the convergence rates for accelerated samplers \citep{luhman2021knowledge,lu2022dpm,zheng2023dpm,zhao2024unipc,zhangfast,jolicoeur2021gotta,xue2024sa,li2024accelerating,huang2024convergence,wu2024stochastic,huang2025fast,li2025faster}. In addition, it remains unclear if we can obtain sharper convergence rates when the target distribution exhibits low-dimensional structure (e.g., low intrinsic dimension \citep{li2024adapting,li2024d, huang2024denoising}). Furthermore, it would be worthwhile to explore if the non-uniform Lipschitz property holds for all 
 distributions on $\mathbb{R}^d$ with $L \lesssim \text{poly}(\log(dT))$. 

\section*{Acknowledgments}
G.~Li is supported in part by the Chinese University of Hong Kong Direct Grant for Research and the Hong Kong Research Grants Council ECS 2191363.

\appendix

\section{Numerical experiments}
\label{app:simulations}

In this section, we conduct numerical experiments to verify our theoretical findings. Specifically, we take the target distribution $p_{\mathsf{data}}$ to be a $d$-dimensional Gaussian distribution with zero mean and independent coordinates: its covariance is diagonal with variances $\{\sigma_i^2\}_{i=1}^d$ drawn i.i.d.~from Unif$[0,10]$. We assume access to the exact score, i.e., $s_t(\cdot) = s_t^{\star}(\cdot)$.
Parameters $\eta_t$ and $\sigma_t^2$ are chosen as \eqref{eq:def-eta-sigma}, and $\{\overline{\alpha}_t\}_{t=1}^T$ (or $\{\beta_t\}_{t=1}^T$) are set as \eqref{eq:learning-rate-example}
with $c_0=2$ and $c_1 = 4$.
Under this choice, each $Y_{t}$ in \eqref{eq:DDPM} is Gaussian, and hence the KL divergence between $X_{1}$ and $Y_1$ has a closed-form expression.

To show the dependence on the number of iterations $T$ and data dimension $d$, 
we implement DDPM \eqref{eq:DDPM} with $d$ fixed at $100$ and the number of iterations $T$ varying from $100$ to $10000$,
and with $T$ fixed at $1000$ and $d$ varying from $10$ to $1000$,
and compute the KL divergence between distributions $p_{X_1}$ and the output $Y_1$.

\begin{figure}
\begin{center}
\includegraphics[width=\textwidth]{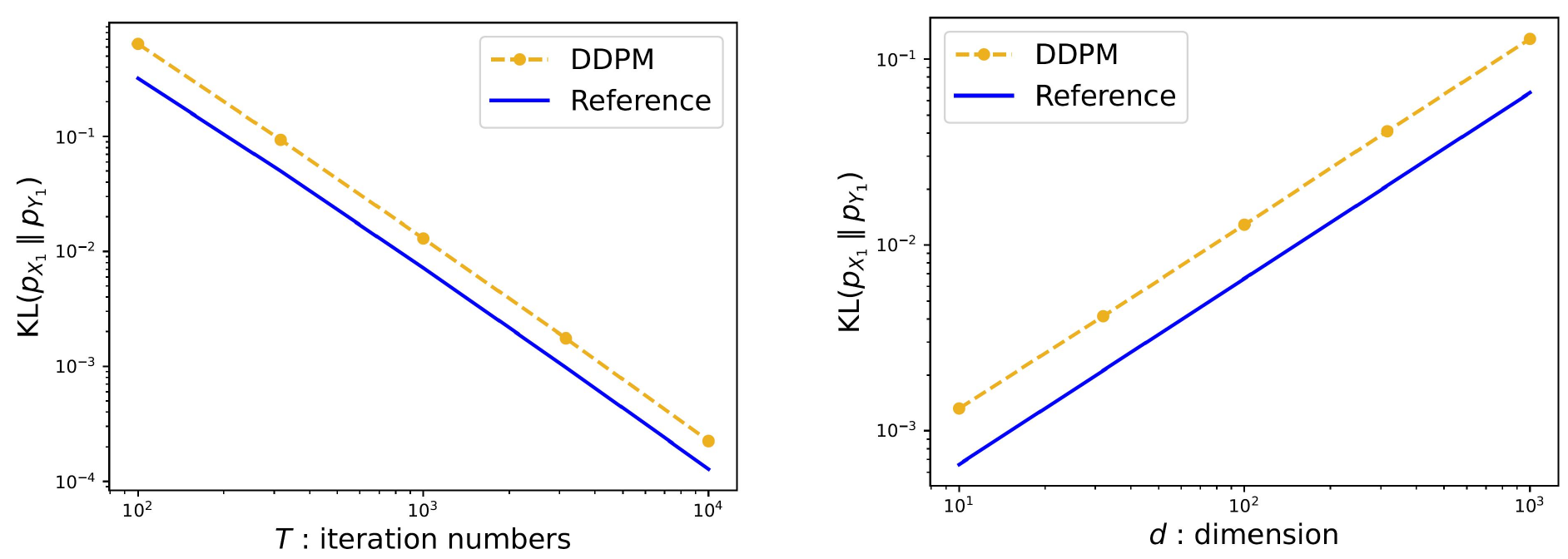}
\end{center}
\caption{KL divergence of DDPM and fitted rate $\Theta(d \log^4 T/T^2)$:
(a) $d = 100$; (b) $T = 1000$. \label{fig:exp}}
\end{figure}

The results are presented in Figure \ref{fig:exp}.
The orange line represents the empirical results, and the blue line corresponds to the theoretical rate $O(\log^2T/T^2)$.
According to Theorem \ref{thm:main}, 
our theoretical analysis predicts a convergence rate of $O(d\log^4 T/T^2)$ in terms of KL divergence in case of Gaussian distribution, 
which is consistent with empirical observations ignoring dependence on logarithmic factors. 
Refining this dependency requires further effort and is left for future work.

\section{Proof of Lemma \ref{lem:ODE}}
\label{sec:proof-lem-ODE}

The proof is similar to \citet[Lemma 2]{li2024improved}; we include it here for completeness.
We shall establish Lemma \ref{lem:ODE} via induction.
From the definition of $\widehat{X}_{T}$, we have $\widehat{X}_{T}\overset{\rm{d}}{=}X_T$.
Assume that $\widehat{X}_{t}\overset{\rm{d}}{=}X_t\overset{\rm{d}}{=}\overline{X}_{\tau_{T-t+1}}$.
We intend to prove that $\widehat{X}_{t-1}\overset{\rm{d}}{=}X_{t-1}$.
In view of \eqref{eq:def-hatX}, we have
\begin{align}
	\frac{\widehat{X}_{t-1}}{\sqrt{\overline{\alpha}_{t-1}}} 
	&= \frac{1}{\sqrt{\overline{\alpha}_t}}\widehat{X}_t + \int_{\widehat{\tau}_{T-t+2}}^{\tau_{T-t+1}}\frac{\overline{s}_\tau^{\star}(x_\tau)}{2(1-\tau)^{\frac32}}\mathrm{d}\tau + \frac{\sigma_t}{\sqrt{\overline{\alpha}_t}} Z_t\notag\\
	&= \frac{1}{\sqrt{1-\tau_{T-t+1}}}\widehat{X}_t + \int_{\widehat{\tau}_{T-t+2}}^{\tau_{T-t+1}}\frac{\overline{s}_\tau^{\star}(x_\tau)}{2(1-\tau)^{\frac32}}\mathrm{d}\tau + \frac{\sigma_t}{\sqrt{\overline{\alpha}_t}} Z_t\notag\\
	&= \frac{1}{\sqrt{1-\widehat{\tau}_{T-t+2}}}\Phi_{\tau_{T-t+1}\to \widehat{\tau}_{T-t+2}}(\widehat{X}_t)  + \frac{\sigma_t}{\sqrt{\overline{\alpha}_t}} Z_t\notag\\
	&\overset{\rm{d}}{=} \frac{X_{\widehat{\tau}_{T-t+2}}}{\sqrt{1-\widehat{\tau}_{T-t+2}}}+ \frac{\sigma_t}{\sqrt{\overline{\alpha}_t}} Z_t\label{eq:proof-lem-ODE-temp-1}\\
	&\overset{\rm{d}}{=}X_0 +  \left(\frac{\widehat{\tau}_{T-t+2}}{1-\widehat{\tau}_{T-t+2}} + \frac{\sigma_t^2}{\overline{\alpha}_t}\right)^{1/2} Z_t',\notag
\end{align}
where $Z_t'\sim\mathcal{N}(0,I_d)$. Here, the penultimate line is due to \eqref{eq:ODE-d}, and the last line makes use of Lemma~\ref{lem:ODE}.
It suffices to prove that
\begin{align*}
	\frac{\widehat{\tau}_{T-t+2}}{1-\widehat{\tau}_{T-t+2}} + \frac{\sigma_t^2}{\overline{\alpha}_t} = \frac{1 - \overline{\alpha}_{t-1}}{\overline{\alpha}_{t-1}},
\end{align*}
which can be completed by direct calculation:
\begin{align*}
	\frac{\widehat{\tau}_{T-t+2}}{1-\widehat{\tau}_{T-t+2}} + \frac{\sigma_t^2}{\overline{\alpha}_t} 
	= \frac{2\alpha_t - 1 - \overline{\alpha}_t}{\overline{\alpha}_t} + \frac{1 - \alpha_t}{\overline{\alpha}_t}
	= \frac{1-\overline{\alpha}_{t-1}}{\overline{\alpha}_{t-1}}.
\end{align*}

\section{Proof of Lemma \ref{lem:discretization}}
\label{sec:proof-lem-discretization}

Before diving into the proof details, we make the observation that
\begin{align}\label{eq:proof-lem-discretization-temp-2}
	\int_{\widetilde{\tau}_{T-t+2}}^{\tau_{T-t+1}}\frac{1}{2(1-\tau)^{\frac 32}\tau^{\frac 12}} \mathrm{d}\tau \overset{\text{(i)}}{=} \sqrt{\frac{\tau_{T-t+1}}{1-\tau_{T-t+1}}} - \sqrt{\frac{\widetilde{\tau}_{T-t+2}}{1-\widetilde{\tau}_{T-t+2}}} \overset{\text{(ii)}}{=}  \sqrt{\frac{1-\overline{\alpha}_t}{\overline{\alpha}_t}} - \sqrt{\frac{1-\widetilde{\alpha}_{t-1}}{\widetilde{\alpha}_{t-1}}},
\end{align}
where (i) holds since
\begin{align}\label{eq:proof-lem-discretization-integral}
	\frac{\mathrm{d} }{\mathrm{d} \tau}\sqrt{\frac{\tau}{1-\tau}} = \frac{1}{2}\sqrt{\frac{1-\tau}{\tau}}\frac{1}{(1-\tau)^2} = \frac{1}{2(1-\tau)^{\frac 32}\tau^{\frac 12}},
\end{align}
and (ii) arises from the definition of $\widetilde{\tau}_{T-t+2}$ in \eqref{eq:def-tildetau}.
Recalling the definition of $\widetilde{\alpha}_{t-1}$ (cf.~\eqref{eq:def-tildetau}), one has
\begin{align*}
	\frac{1-\widetilde{\alpha}_{t-1}}{\widetilde{\alpha}_{t-1}} 
	&= \frac{\overline{\alpha}_{t-1}(1-\alpha_t) + \alpha_t(1-\overline{\alpha}_{t-1})-\overline{\alpha}_{t-1}(1-\overline{\alpha}_t)}{\overline{\alpha}_{t-1}(1-\overline{\alpha}_t)} \notag\\
	&= \frac{\alpha_t - 2\overline{\alpha}_t + \overline{\alpha}_t\overline{\alpha}_{t-1}}{\overline{\alpha}_{t-1}(1-\overline{\alpha}_t)} 
	= \frac{\alpha_t(1-\overline{\alpha}_{t-1})^2}{\overline{\alpha}_{t-1}(1-\overline{\alpha}_t)} = \frac{\alpha_t^2(1-\overline{\alpha}_{t-1})^2}{\overline{\alpha}_{t}(1-\overline{\alpha}_t)} = \frac{(\alpha_t-\overline{\alpha}_{t})^2}{\overline{\alpha}_{t}(1-\overline{\alpha}_t)}.
\end{align*}
Combining \eqref{eq:proof-lem-discretization-temp-2} and the previous equation, we obtain
\begin{align*}
	\int_{\widetilde{\tau}_{T-t+2}}^{\tau_{T-t+1}}\frac{1}{2(1-\tau)^{\frac 32}\tau^{\frac 12}} \mathrm{d}\tau = \sqrt{\frac{1-\overline{\alpha}_t}{\overline{\alpha}_t}} - \sqrt{\frac{1-\widetilde{\alpha}_{t-1}}{\widetilde{\alpha}_{t-1}}} 
    = \sqrt{\frac{1-\overline{\alpha}_t}{\overline{\alpha}_t}} - \frac{\alpha_t-\overline{\alpha}_{t}}{\sqrt{\overline{\alpha}_{t}(1-\overline{\alpha}_t)}}
	= \frac{1-\alpha_t}{\sqrt{\overline{\alpha}_{t}(1-\overline{\alpha}_t)}}, 
\end{align*}
Consequently, \(\eta_t s_t^{\star}(x)\) can be expressed as the integral
\begin{align}\label{eq:formulation-st}
	\eta_ts_t^{\star}(x) 
	&= \eta_t\overline{s}_{\tau_{T-t+1}}^{\star}(x) =\frac{\eta_t}{\sqrt{\overline{\alpha}_{t}(1-\overline{\alpha}_t)}}\sqrt{\overline{\alpha}_t\tau_{T-t+1}}\overline{s}_{\tau_{T-t+1}}^{\star}(x)\notag\\
	&= 
	\sqrt{\overline{\alpha}_t}\int_{\widetilde{\tau}_{T-t+2}}^{\tau_{T-t+1}} \frac{\sqrt{\tau_{T-t+1}}\overline{s}_{\tau_{T-t+1}}^{\star}(x)}{2(1-\tau)^{\frac 32}\tau^{\frac 12}} \mathrm{d}\tau.
\end{align}

Now we are ready to bound the expectation of $\|\xi_{t-1}(x_t)\|_2^2$. Recalling the definition of $\mu_{\widehat{X}_{t-1}|\widehat{X}_t}(x_t)$, we have
\begin{align}\label{eq:proof-lem-discretization-temp-1}
	\xi_{t-1}(x_t) 
	&=\frac{\sqrt{\alpha_t}}{\sigma_t}\left(\frac{\sqrt{\overline{\alpha}_{t-1}}}{\sqrt{1-\widehat{\tau}_{T-t+2}}}\Phi_{\tau_{T-t+1}\to\widehat{\tau}_{T-t+2}}(x_t)-\frac{1}{\sqrt{\alpha_t}}(x_t + \eta_t s_t^{\star}(x_t))\right)\notag\\
	&=\frac{1}{\sigma_t}\left(x_t + \sqrt{\alpha_t}\int_{\widehat{\tau}_{T-t+2}}^{\tau_{T-t+1}}\frac{\sqrt{\overline{\alpha}_{t-1}}\overline{s}_\tau^{\star}(x_\tau)}{2(1-\tau)^{\frac32}}\mathrm{d}\tau-x_t - \eta_t s_t^{\star}(x_t)\right)\notag\\
	&=\frac{1}{\sigma_t}\left(\sqrt{\overline{\alpha}_{t}}\int_{\widehat{\tau}_{T-t+2}}^{\tau_{T-t+1}}\frac{\overline{s}_\tau^{\star}(x_\tau)}{2(1-\tau)^{\frac32}}\mathrm{d}\tau- \eta_t s_t^{\star}(x_t)\right).
\end{align}
We introduce the following auxiliary variable:
\begin{align}\label{eq:def-zetat}
\zeta_t(x_t)\coloneqq \frac{\sqrt{\overline{\alpha}_t}}{\sigma_t}\int_{\widehat{\tau}_{T-t+2}}^{\widetilde{\tau}_{T-t+2}}\frac{\overline{s}_\tau^{\star}(x_\tau)}{2(1-\tau)^{\frac32}}\mathrm{d}\tau.
\end{align}
Putting \eqref{eq:formulation-st} and \eqref{eq:proof-lem-discretization-temp-1} together, it follows that
\begin{align}
	\xi_{t-1}(x_t) &=\frac{\sqrt{\overline{\alpha}_{t}}}{\sigma_t}\left(\int_{\widetilde{\tau}_{T-t+2}}^{\tau_{T-t+1}}\frac{\overline{s}_\tau^{\star}(x_\tau)}{2(1-\tau)^{\frac32}}\mathrm{d}\tau- \int_{\widetilde{\tau}_{T-t+2}}^{\tau_{T-t+1}} \frac{\sqrt{\tau_{T-t+1}}\overline{s}_{\tau_{T-t+1}}^{\star}(x_{t})}{2(1-\tau)^{\frac 32}\tau^{\frac 12}} \mathrm{d}\tau\right) + \zeta_t(x_t)\notag\\
	&=\frac{\sqrt{\overline{\alpha}_{t}}}{\sigma_t}\int_{\widetilde{\tau}_{T-t+2}}^{\tau_{T-t+1}}\frac{1}{2(1-\tau)^{\frac32}\tau^{\frac 12}}\left(\sqrt{\tau}\overline{s}_\tau^{\star}(x_\tau) - \sqrt{\tau_{T-t+1}}\overline{s}_{\tau_{T-t+1}}^{\star}(x_{t})\right)\mathrm{d}\tau + \zeta_t(x_t)\notag\\
	&=-\frac{\sqrt{\overline{\alpha}_{t}}}{\sigma_t}\int_{\widetilde{\tau}_{T-t+2}}^{\tau_{T-t+1}}\frac{1}{2(1-\tau)^{\frac32}\tau^{\frac 12}}\int_\tau^{\tau_{T-t+1}}\frac{\partial \sqrt{\tau'}\overline{s}_{\tau'}^{\star}(x_\tau')}{\partial \tau'} \mathrm{d}\tau' \mathrm{d}\tau + \zeta_t(x_t)\notag\\
	&=-\frac{\sqrt{\overline{\alpha}_{t}}}{\sigma_t}\int_{\widetilde{\tau}_{T-t+2}}^{\tau_{T-t+1}}\frac{\partial \sqrt{\tau'}\overline{s}_{\tau'}^{\star}(x_\tau')}{\partial \tau'}\int_{\widetilde{\tau}_{T-t+2}}^{\tau'} \frac{1}{2(1-\tau)^{\frac32}\tau^{\frac 12}}\mathrm{d}\tau \mathrm{d}\tau' + \zeta_t(x_t)\notag\\
	&=\frac{\sqrt{\overline{\alpha}_{t}}}{\sigma_t}\int_{\widetilde{\tau}_{T-t+2}}^{\tau_{T-t+1}}\left( \sqrt{\frac{\widetilde{\tau}_{T-t+2}}{1-\widetilde{\tau}_{T-t+2}}}-\sqrt{\frac{\tau'}{1-\tau'}}\right)\frac{\partial \sqrt{\tau'}\overline{s}_{\tau'}^{\star}(x_\tau')}{\partial \tau'} \mathrm{d}\tau' + \zeta_t(x_t),
\end{align}
where the third line makes use of fundamental theorem of calculus, the fourth line invokes follows by exchanging the order of integration, and the last equation holds due to \eqref{eq:proof-lem-discretization-integral}.
Therefore, one can bound the norm of $\xi_{t-1}(x_t) - \zeta_t(x_t)$ as follows
\begin{align}\label{eq:proof-lem-discretization-temp-4}
	\|\xi_{t-1}(x_t)-\zeta_t(x_t)\|_2&\le \left| \sqrt{\frac{\widetilde{\tau}_{T-t+2}}{1-\widetilde{\tau}_{T-t+2}}}-\sqrt{\frac{\tau_{T-t+1}}{1-\tau_{T-t+1}}}\right|\frac{\sqrt{\overline{\alpha}_{t}}}{\sigma_t}\int^{\tau_{T-t+1}}_{\widetilde{\tau}_{T-t+2}}\left\|\frac{\partial \sqrt{\tau'}\overline{s}_{\tau'}^{\star}(x_\tau')}{\partial \tau'}\right\|_2 \mathrm{d}\tau'.
\end{align}
In the rest of the proof, we bound $\|\xi_{t-1}(x_t)\|_2^2$ using the following lemma, which establishes key properties of $\widetilde{\tau}_{T-t+2}$.
Its proof is postponed to Appendix \ref{subsec:proof-lem-tildetau}.
\begin{lemma}\label{lem:tildetau}
	For any $1\le t\le T$, the $\widetilde{\tau}_{T-t+2}$ and $\widetilde{\alpha}_{t-1}$ defined in \eqref{eq:def-hattau} satisfy
	\begin{subequations}
		\begin{align}
			&\widetilde{\tau}_{T-t+2}  \le {\tau}_{T-t+2},\qquad {\tau}_{T-t+1} - \widetilde{\tau}_{T-t+2}=\widetilde{\alpha}_{t-1} - \overline{\alpha}_{t}\lesssim \frac{\log T}{T}\overline{\alpha}_{t-1}(1-\overline{\alpha}_{t}),\label{eq:lem-tildetau-1}\\
			&\frac{1-\overline{\alpha}_t}{1-\widetilde{\alpha}_{t-1}} \lesssim 1,\label{eq:lem-tildetau-2}\\
			&\left|\sqrt{\frac{\tau_{T-t+1}}{1-\tau_{T-t+1}}} - \sqrt{\frac{\widetilde{\tau}_{T-t+2}}{1-\widetilde{\tau}_{T-t+2}}}\right|\lesssim \frac{\log T}{T}\sqrt{\frac{1 - \overline{\alpha}_{t}}{\overline{\alpha}_{t}}},\label{eq:lem-tildetau-3}\\
            &\mathbb{E}[\|\zeta_t(x_t)\|_2^2]\lesssim \frac{d\log^3T}{T^3}.\label{eq:lem-tildetau-4}
		\end{align}    
	\end{subequations}
\end{lemma}


Inserting \eqref{eq:lem-tildetau-3} into \eqref{eq:proof-lem-discretization-temp-4} yields
\begin{align*}
    \|\xi_{t-1}(x_t)-\zeta_t(x_t)\|_2&\le \frac{2c\log T}{T}\frac{\sqrt{1-\overline{\alpha}_{t}}}{\sigma_t}\int_{\widetilde{\tau}_{T-t+2}}^{\tau_{T-t+1}}\left\|\frac{\partial \sqrt{\tau'}\overline{s}_{\tau'}^{\star}(x_\tau')}{\partial \tau'}\right\|_2 \mathrm{d}\tau'.
\end{align*}
Applying Cauchy-Schwarz inequality leads to
\begin{align}\label{ineq:norm_xi_t}
	 \|\xi_{t-1}(x_t)-\zeta_t(x_t)\|_2^2
    &\le\frac{4c^2\log^2 T}{T^2}\frac{1-\overline{\alpha}_{t}}{\sigma_t^2}(\tau_{T-t+1}-\widetilde{\tau}_{T-t+2})\int_{\widetilde{\tau}_{T-t+2}}^{\tau_{T-t+1}}\left\|\frac{\partial \sqrt{\tau}\overline{s}_{\tau}^{\star}(x_\tau)}{\partial \tau}\right\|_2^2 \mathrm{d}\tau\notag\\
	&\overset{\text{(i)}}{\lesssim} \frac{c^3\log^3 T}{T^3}\frac{\overline{\alpha}_{t-1}(1-\overline{\alpha}_{t})^2}{\sigma_t^2}\int_{\widetilde{\tau}_{T-t+2}}^{\tau_{T-t+1}}\left\|\frac{\partial \sqrt{\tau}\overline{s}_{\tau}^{\star}(x_\tau)}{\partial \tau}\right\|_2^2 \mathrm{d}\tau\notag\\
	&\overset{\text{(ii)}}{\asymp}\frac{c^3\log^3 T}{T^3}\frac{\overline{\alpha}_{t-1}(1-\overline{\alpha}_{t})^3}{(1-\alpha_t)(\alpha_t - \overline{\alpha}_t)}\int_{\widetilde{\tau}_{T-t+2}}^{\tau_{T-t+1}}\left\|\frac{\partial \sqrt{\tau}\overline{s}_{\tau}^{\star}(x_\tau)}{\partial \tau}\right\|_2^2 \mathrm{d}\tau\notag\\
	&\overset{\text{(iii)}}{\lesssim}\frac{c^2\log^2 T}{T^2}\overline{\alpha}_t(1-\overline{\alpha}_t)\int_{\widetilde{\tau}_{T-t+2}}^{\tau_{T-t+1}}\left\|\frac{\partial \sqrt{\tau}\overline{s}_{\tau}^{\star}(x_\tau)}{\partial \tau}\right\|_2^2 \mathrm{d}\tau,
\end{align}
where (i) arises from \eqref{eq:lem-tildetau-1}, (ii) holds due to the definition of $\sigma_t^2$ in \eqref{eq:def-eta-sigma}, and (iii) results from
\begin{align}\label{ineq:alpha_t_property}
	\frac{\overline{\alpha}_{t-1}}{\overline{\alpha}_{t}} \stackrel{\eqref{eq:learning-rate}}{\le} 1+\frac{c\log T}{T}(1 - \overline{\alpha}_{t}) \leq 1+\frac{c\log T}{T},~\quad~\frac{1 - \overline{\alpha}_{t-1}}{1 - \overline{\alpha}_{t}}  \stackrel{\eqref{eq:learning-rate}}{\ge} 1 - \frac{c\log T}{T}\overline{\alpha}_{t} \geq 1 - \frac{c\log T}{T}
\end{align}
and
\begin{align*}
	\frac{\overline{\alpha}_{t-1}(1-\overline{\alpha}_{t})^3}{(1-\alpha_t)(\alpha_t - \overline{\alpha}_t)} 
	&= \frac{\overline{\alpha}_{t-1}^3(1-\overline{\alpha}_{t})^3}{\overline{\alpha}_t(\overline{\alpha}_{t-1}-\overline{\alpha}_t)(1 - \overline{\alpha}_{t-1})}\\ &\stackrel{\eqref{eq:learning-rate}}{=} \frac{T\overline{\alpha}_{t-1}^3(1-\overline{\alpha}_{t})^3}{c\log T\overline{\alpha}_t^2(1-\overline{\alpha}_t)(1 - \overline{\alpha}_{t-1})}\\ 
	&= \frac{T\overline{\alpha}_{t}(1-\overline{\alpha}_{t})}{c\log T}\left(\frac{\overline{\alpha}_{t-1}}{\overline{\alpha}_{t}}\right)^3\frac{1 - \overline{\alpha}_{t}}{1 - \overline{\alpha}_{t-1}}\\ &\stackrel{\eqref{ineq:alpha_t_property}}{\le} \frac{T\overline{\alpha}_{t}(1-\overline{\alpha}_{t})}{c\log T}\left(1+\frac{c\log T}{T}\right)^3\frac{1}{1-\frac{c\log T}{T}}\\ &\le \frac{2T\overline{\alpha}_{t}(1-\overline{\alpha}_{t})}{c\log T},
\end{align*}
provided that $T\ge 6c\log T$.
Taking expectation with respect to $x_t$ in \eqref{ineq:norm_xi_t} and applying Lemma~\ref{lem:ODE} together with \eqref{eq:ODE-d}, 
we obtain
\begin{align}\label{ineq:norm_xi_t_expectation}
	\mathbb{E}_{x_t\sim\widehat{X}_t}\left[\|\xi_{t-1}(x_t)-\zeta_t(x_t)\|_2^2\right]
	&\lesssim \frac{c^2\log^2 T}{T^2}\overline{\alpha}_{t}(1-\overline{\alpha}_{t})
	\int_{\widetilde{\tau}_{T-t+2}}^{\tau_{T-t+1}}\mathbb{E}_{x_\tau\sim\overline{X}_\tau}\left[\left\|\frac{\partial \sqrt{\tau}\overline{s}_{\tau}^{\star}(x_\tau)}{\partial \tau}\right\|_2^2\right] \mathrm{d}\tau + \frac{d\log^3T}{T^3}.
\end{align}
Furthermore, the following lemma bounds the right-hand side of \eqref{ineq:norm_xi_t_expectation}; its proof is deferred to Appendix \ref{subsec:proof-lem-bound-derivative}.
\begin{lemma}\label{lem:bound-derivative}
	For any $0<\tau<1$, we have
	\begin{align}
		\mathbb{E}_{x_\tau\sim\overline{X}_\tau}\left[\left\|\frac{\partial \sqrt{\tau}\overline{s}_{\tau}^{\star}(x_\tau)}{\partial \tau}\right\|_2^2\right]\lesssim \frac{d}{\tau^2(1-\tau)^2}\left(\log T \mathbb{E}_{x_\tau\sim \overline{X}_\tau} \left\|\Sigma_\tau(x_\tau)\right\|_{\mathsf{op}}^2+ 1\right) +\frac{d\min\left\{d\log T,L\right\}}{\tau^2(1-\tau)^2}.
	\end{align}
\end{lemma}
By virtue of Lemma~\ref{lem:bound-derivative}, \eqref{eq:lem-tildetau-3} and \eqref{ineq:norm_xi_t_expectation}, one has
\begin{align*}
	\mathbb{E}_{x_t\sim\widehat{X}_t}\left[\|\xi_{t-1}(x_t)\|_2^2\right] &\leq 2\left(\mathbb{E}_{x_t\sim\widehat{X}_t}\left[\|\xi_{t-1}(x_t) - \|\zeta_t(x_t)\|_2^2\|_2^2\right] + \mathbb{E}_{x_t\sim\widehat{X}_t}\left[\|\|\zeta_t(x_t)\|_2^2\|_2^2\right]\right)\notag\\
	&\lesssim \frac{d\overline{\alpha}_{t}(1-\overline{\alpha}_{t})\log^3 T}{T^2\widetilde{\tau}_{T-t+2}^2}
	\int_{\widetilde{\tau}_{T-t+2}}^{\tau_{T-t+1}}\frac{\mathbb{E}_{x_\tau\sim \overline{X}_\tau} \left\|\Sigma_\tau(x_\tau)\right\|_{\mathsf{op}}^2}{(1-\tau)^2} \mathrm{d} \tau\notag\\
	&\qquad +\frac{d\min\left\{d\log T,L\right\}\overline{\alpha}_{t}(1-\overline{\alpha}_{t})\log^2 T}{T^2\widetilde{\tau}_{T-t+2}^2(1-\tau_{T-t+1})^2}(\tau_{T-t+1}-\widetilde{\tau}_{T-t+2}) + \frac{d\log^3T}{T^3}\notag\\
	&\lesssim \frac{d\overline{\alpha}_{t}\log^3 T}{T^2(1-\overline{\alpha}_{t})}\int_{\widetilde{\tau}_{T-t+2}}^{\tau_{T-t+1}}\frac{\mathbb{E}_{x_\tau\sim \overline{X}_\tau} \left\|\Sigma_\tau(x_\tau)\right\|_{\mathsf{op}}^2}{(1-\tau)^2} \mathrm{d} \tau + 
	\frac{d\min\left\{d\log T,L\right\}\log^3 T}{T^3},
\end{align*}
where the last inequality makes use of the following facts:
\begin{align*}
	\frac{\overline{\alpha}_{t}(1-\overline{\alpha}_{t})}{T^2\widetilde{\tau}_{T-t+2}^2}&\stackrel{\eqref{eq:lem-tildetau-1}}{\le}\frac{\overline{\alpha}_{t}}{T^2(1-\overline{\alpha}_{t})}\frac{(1-\overline{\alpha}_{t})^2}{(1-\overline{\alpha}_{t-2})^2}\lesssim \frac{\overline{\alpha}_{t}}{T^2(1-\overline{\alpha}_{t})},
	\notag\\
	\frac{\overline{\alpha}_{t}(1-\overline{\alpha}_{t})}{T^2\widetilde{\tau}_{T-t+2}^2(1-\tau_{T-t+1})^2}(\tau_{T-t+1}-\widetilde{\tau}_{T-t+2}) &\stackrel{\eqref{eq:lem-tildetau-1}}{\lesssim} \frac{\log T}{T^3} \frac{\overline{\alpha}_{t-1}}{\overline{\alpha}_{t}} \frac{(1-\overline{\alpha}_{t})^2}{(1-\overline{\alpha}_{t-2})^2}\lesssim \frac{\log T}{T^2}.
\end{align*}

\subsection{Proof of Lemma \ref{lem:tildetau}}
\label{subsec:proof-lem-tildetau}
Recalling the definitions of $\widetilde{\tau}_{T-t+2}$ and $\widetilde{\alpha}_{t-1}$ (cf.~\eqref{eq:def-tildetau}), we have
\begin{align}\label{ineq:alpha_tilde_and_alpha}
	\widetilde{\tau}_{T-t+2} -  {\tau}_{T-t+2}
	& = \overline{\alpha}_{t-1} - \widetilde{\alpha}_{t-1}
	= \overline{\alpha}_{t-1} - \frac{\overline{\alpha}_{t-1}(1-\overline{\alpha}_t)}{\overline{\alpha}_{t-1}(1-\alpha_t) + \alpha_t(1-\overline{\alpha}_{t-1})}\notag\\
    &=\overline{\alpha}_{t-1} \frac{\overline{\alpha}_{t-1}(1-\alpha_t) + \alpha_t(1-\overline{\alpha}_{t-1}) - (1-\overline{\alpha}_t)}{\overline{\alpha}_{t-1}(1-\alpha_t) + \alpha_t(1-\overline{\alpha}_{t-1})}\le 0,
\end{align}
where the last inequality arises from 
$$
\overline{\alpha}_{t-1}(1-\alpha_t) + \alpha_t(1-\overline{\alpha}_{t-1}) - (1-\overline{\alpha}_t) = \overline{\alpha}_{t-1} + \alpha_t - \overline{\alpha}_{t}-1 = (\overline{\alpha}_{t-1}-1)(1-\alpha_t)\le 0.
$$

Furthermore, one can show that
\begin{align*}
	{\tau}_{T-t+1} - \widetilde{\tau}_{T-t+2} &\le {\tau}_{T-t+1} - {\tau}_{T-t+3} = \overline{\alpha}_{t-2} - \overline{\alpha}_{t} = (\overline{\alpha}_{t-1} - \overline{\alpha}_{t}) + (\overline{\alpha}_{t-2} - \overline{\alpha}_{t-1})\notag\\
	&\stackrel{\eqref{eq:learning-rate}}{\le} \frac{c\log T}{T}\overline{\alpha}_{t}(1-\overline{\alpha}_{t}) + \frac{c\log T}{T}\overline{\alpha}_{t-1}(1-\overline{\alpha}_{t-1})\le \frac{2c\log T}{T}\overline{\alpha}_{t-1}(1-\overline{\alpha}_{t}),
\end{align*}
which has finished the proof of the last inequality in \eqref{eq:lem-tildetau-1}.

%
%
Next, we verify \eqref{eq:lem-tildetau-2} via a direct calculation:
\begin{align*}
	\frac{1-\overline{\alpha}_t}{1-\widetilde{\alpha}_{t-1}} 
	&\stackrel{\eqref{eq:def-hattau}}{=} (1-\overline{\alpha}_t)\frac{\overline{\alpha}_{t-1}(1-\alpha_t) + \alpha_t(1-\overline{\alpha}_{t-1})}{\alpha_t(1-\overline{\alpha}_{t-1})^2}
	= \frac{1-\overline{\alpha}_t}{1-\overline{\alpha}_{t-1}}\frac{\overline{\alpha}_{t-1}(\overline{\alpha}_{t-1}-\overline{\alpha}_t) + \overline{\alpha}_t(1-\overline{\alpha}_{t-1})}{\overline{\alpha}_t(1-\overline{\alpha}_{t-1})}\notag\\
	&\stackrel{\eqref{eq:learning-rate}}{\le
    }\frac{1}{1-\frac{c\overline{\alpha}_t\log T}{T}}\left(1 + \frac{c\overline{\alpha}_{t-1}\log T}{T}\frac{1-\overline{\alpha}_{t}}{1-\overline{\alpha}_{t-1}}\right)\le \frac{1}{1-\frac{c\log T}{T}}\left(1 + \frac{2c\log T}{T}\right) \le 2,
\end{align*}
as long as $T\ge 4c\log T$, where the last inequality makes use of the fact that
\begin{align}\label{eq:ratio-1-alpha}
	\frac{1-\overline{\alpha}_{t}}{1-\overline{\alpha}_{t-1}} = 1 + \frac{\overline{\alpha}_{t-1}-\overline{\alpha}_{t}}{1-\overline{\alpha}_{t-1}} \le 1 + \frac{c\overline{\alpha}_t\log T}{T}\frac{1-\overline{\alpha}_t}{1-\overline{\alpha}_{t-1}}  \le \frac{1}{1 - \frac{c\overline{\alpha}_t\log T}{T}}\le \frac{4}{3}.
\end{align}
Now we turn to establishing \eqref{eq:lem-tildetau-3}. In view of \eqref{eq:proof-lem-discretization-integral}, we have
\begin{align*}
	\left|\sqrt{\frac{\tau_{T-t+1}}{1-\tau_{T-t+1}}} - \sqrt{\frac{\widetilde{\tau}_{T-t+2}}{1-\widetilde{\tau}_{T-t+2}}}\right|
	& = \int_{\widetilde{\tau}_{T-t+2}}^{\tau_{T-t+1}} \frac{1}{2(1-\tau)^{\frac 32}\tau^{\frac 12}}{\rm d}\tau \le \frac{\tau_{T-t+1}-\widetilde{\tau}_{T-t+2}}{2(1-\tau_{T-t+1})^{\frac 32}\widetilde{\tau}_{T-t+2}^{\frac 12}} = \frac{\overline{\alpha}_t - \widetilde{\alpha}_{t-1}}{2\overline{\alpha}_{t}^{\frac 32}(1-\widetilde{\alpha}_{t-1})^{\frac 12}}\notag\\
	&\overset{\text{(i)}}{\le} \frac{c\log T}{T}\frac{\overline{\alpha}_{t-1}(1-\overline{\alpha}_{t})}{\overline{\alpha}_{t}^{\frac 32}(1-\widetilde{\alpha}_{t-1})^{\frac 12}}\overset{\text{(ii)}}{\le} \frac{2c\log T}{T}\sqrt{\frac{1 - \overline{\alpha}_{t}}{\overline{\alpha}_{t}}},
\end{align*}
where (i) invokes \eqref{eq:lem-tildetau-1}, and (ii) arises from \eqref{eq:ratio-1-alpha} and 
\begin{align*}
	\frac{\overline{\alpha}_{t-1}}{\overline{\alpha}_{t}}\sqrt{\frac{1-\overline{\alpha}_{t}}{1-\widetilde{\alpha}_{t-1}}}  \stackrel{\eqref{eq:learning-rate}~\text{and}~\eqref{eq:lem-tildetau-2}}{\le} \left(1 + \frac{c\log T}{T}\right)\sqrt{2} \le 2,
\end{align*}
provided that $T\ge 4c\log T$.

Finally, we show that \eqref{eq:lem-tildetau-4} holds. Eqn.~\eqref{eq:learning-rate} tells us that
\begin{align}
	1 - \alpha_t &\leq \frac{c\log T}{T}\alpha_t(1 - \overline{\alpha}_t) \leq \frac{c\log T}{T} \leq \frac{1}{4},\label{ineq:property_alpha_t_1}\\
	1 - \overline{\alpha}_{t-1} &\geq 1 - \overline{\alpha}_{t} - \frac{c\log T}{T}\overline{\alpha}_t(1-\overline{\alpha}_t) \geq \left(1 - \frac{c\log T}{T}\right)\left(1 - \overline{\alpha}_{t}\right) \geq \frac{1}{2}\left(1 - \overline{\alpha}_{t}\right).\label{ineq:property_alpha_t_2}
\end{align}
We make the observation that
\begin{align}\label{ineq:difference_tau_hat_tilde}
	\widetilde{\tau}_{T-t+2} - \widehat{\tau}_{T-t+2} &= \widehat{\alpha}_{t-1} - \widetilde{\alpha}_{t-1}\notag\\ &= \frac{ \overline{\alpha}_{t}}{2\alpha_t-1}  - \frac{\overline{\alpha}_{t-1}(1-\overline{\alpha}_t)}{\overline{\alpha}_{t-1}(1-\alpha_t) + \alpha_t(1-\overline{\alpha}_{t-1})}\notag\\ &= \frac{\overline{\alpha}_{t-1}(1-\alpha_t)^2}{(2\alpha_t-1)(\overline{\alpha}_{t-1}(1-\alpha_t) + \alpha_t(1-\overline{\alpha}_{t-1}))}\notag\\
	&\stackrel{\eqref{ineq:property_alpha_t_1}}{\geq} 0.
\end{align}
Applying Cauchy-Schwarz inequality yields
\begin{align*}
	\|\zeta_t(x_t)\|_2^2 &\leq \frac{\overline{\alpha}_t}{\sigma_t^2}\int_{\widehat{\tau}_{T-t+2}}^{\widetilde{\tau}_{T-t+2}}\frac{1}{4\tau(1 - \tau)^3}\mathrm{d}\tau\int_{\widehat{\tau}_{T-t+2}}^{\widetilde{\tau}_{T-t+2}}\tau \|\overline{s}_\tau^{\star}(x_\tau)\|_2^2\mathrm{d}\tau\notag\\
	&\leq \frac{\overline{\alpha}_t}{1-\alpha_t}\frac{\widetilde{\tau}_{T-t+2}-\widehat{\tau}_{T-t+2}}{4\widehat{\tau}_{T-t+2}(1-\widetilde{\tau}_{T-t+2})^3}\int_{\widehat{\tau}_{T-t+2}}^{\widetilde{\tau}_{T-t+2}}\tau \|\overline{s}_\tau^{\star}(x_\tau)\|_2^2\mathrm{d}\tau.
\end{align*}
As a consequence, we have
\begin{align}\label{ineq400}
\mathbb{E}_{x_t \sim X_t}[\|\zeta_t(x_t)\|_2^2]
&\le \frac{\overline{\alpha}_t}{1-\alpha_t}\frac{\widetilde{\tau}_{T-t+2}-\widehat{\tau}_{T-t+2}}{4\widehat{\tau}_{T-t+2}(1-\widetilde{\tau}_{T-t+2})^3}\int_{\widehat{\tau}_{T-t+2}}^{\widetilde{\tau}_{T-t+2}}\tau \mathbb{E}_{x_\tau \sim \overline{X}_\tau}[\|\overline{s}_\tau^{\star}(x_\tau)\|_2^2]\mathrm{d}\tau\notag\\
&\stackrel{\eqref{ineq:technical_2}}{\leq} \frac{\overline{\alpha}_t}{1-\alpha_t}\frac{\widetilde{\tau}_{T-t+2}-\widehat{\tau}_{T-t+2}}{4\widehat{\tau}_{T-t+2}(1-\widetilde{\tau}_{T-t+2})^3}\int_{\widehat{\tau}_{T-t+2}}^{\widetilde{\tau}_{T-t+2}}\tau \left(\frac{d^4}{\tau^4}\right)^{1/4}\mathrm{d}\tau\notag\\
&= \frac{\overline{\alpha}_t}{1-\alpha_t}\frac{(\widetilde{\tau}_{T-t+2}-\widehat{\tau}_{T-t+2})^2d}{4\widehat{\tau}_{T-t+2}(1-\widetilde{\tau}_{T-t+2})^3}\notag\\
&= \frac{\overline{\alpha}_t}{1-\alpha_t}\frac{(\widetilde{\alpha}_{t-1}-\widehat{\alpha}_{t-1})^2d}{4\widetilde{\alpha}_{t-1}^3(1-\widehat{\alpha}_{t-1})}.
\end{align}
In view of \eqref{ineq:difference_tau_hat_tilde}, one has
\begin{align}\label{ineq401}
\widehat{\alpha}_{t-1} - \widetilde{\alpha}_{t-1} & = \frac{\overline{\alpha}_{t-1}(1-\alpha_t)^2}{(2\alpha_t-1)(\overline{\alpha}_{t-1}(1-\alpha_t) + \alpha_t(1-\overline{\alpha}_{t-1}))}\notag\\
&\leq \frac{c\frac{\log T}{T}\overline{\alpha}_t(1 - \overline{\alpha}_t)}{\frac{1}{2}\alpha_t(1-\overline{\alpha}_{t-1})}(1-\alpha_t)\notag\\
&\lesssim \frac{\log T}{T}\overline{\alpha}_{t-1}(1-{\alpha}_t),
\end{align}
where the second line arises from \eqref{eq:learning-rate} and \eqref{ineq:property_alpha_t_1} and the last line makes use \eqref{ineq:property_alpha_t_2}. Moreover, we can verify that
\begin{align}\label{ineq402}
	1-\widehat{\alpha}_{t-1} &= 1 - \frac{\overline{\alpha}_t}{2\alpha_t - 1} = 1 - \frac{\overline{\alpha}_t\cdot \overline{\alpha}_{t-1}}{\overline{\alpha}_t - (\overline{\alpha}_{t-1} - \overline{\alpha}_t)} \stackrel{\eqref{eq:learning-rate}}{\geq} 1 - \frac{\overline{\alpha}_t\cdot \overline{\alpha}_{t-1}}{\overline{\alpha}_t - \frac{c\log T}{T}\overline{\alpha}_t(1 - \overline{\alpha}_t)} = 1 - \frac{\overline{\alpha}_{t-1}}{1 - \frac{c\log T}{T}(1 - \overline{\alpha}_{t})}\notag\\
	&\geq 1 - \frac{\overline{\alpha}_{t-1}}{1 - \frac{c\log T}{T}(1 - \overline{\alpha}_{t-1})} = (1 - \overline{\alpha}_{t-1})\left(1 - \frac{c\log T}{T}\right)\frac{1}{1 - \frac{c\log T}{T}(1 - \overline{\alpha}_{t-1})}\notag\\ &\geq \frac{1-\overline{\alpha}_{t-1}}{4} \geq \frac{1 - \overline{\alpha}_t}{8}.
\end{align}
and
\begin{align}\label{ineq403}
	\widetilde{\alpha}_{t-1} \stackrel{\eqref{ineq:alpha_tilde_and_alpha}}{\geq} \overline{\alpha}_{t-1}.
\end{align}
Putting \eqref{ineq400} - \eqref{ineq403} together, one arrives at
\begin{align*}
	\mathbb{E}[\|\zeta_t(x_t)\|_2^2] &\lesssim \frac{\overline{\alpha}_t}{1-\alpha_t}\frac{\left(\frac{\log T}{T}\overline{\alpha}_{t-1}(1-{\alpha}_t)\right)^2d}{\overline{\alpha}_{t-1}^3(1 - \overline{\alpha}_t)}\\
	&= d\frac{\log^2 T}{T^2}\alpha_t\frac{\overline{\alpha}_{t-1} - \overline{\alpha}_{t}}{\overline{\alpha}_{t-1}(1 - \overline{\alpha}_t)}\\
	&\stackrel{\eqref{eq:learning-rate}}{\lesssim}  d\frac{\log^2 T}{T^2}\alpha_t\frac{\frac{\log T}{T}\overline{\alpha}_{t}(1 - \overline{\alpha}_t)}{\overline{\alpha}_{t-1}(1 - \overline{\alpha}_t)}\\
	&\leq \frac{d\log^3T}{T^3}.
\end{align*}

\subsection{Proof of Lemma \ref{lem:bound-derivative}}
\label{subsec:proof-lem-bound-derivative}
To begin with, let us write
\begin{align}\label{eq:proof-lem-bound-derivatives-temp-1}
	\frac{\partial \sqrt{\tau}\overline{s}_{\tau}^{\star}(x_\tau)}{\partial \tau} 
	&= \frac{\partial}{\partial y}\sqrt{\tau}\overline{s}_\tau^{\star}(\sqrt{1-\tau}y)\big|_{y=x_\tau/\sqrt{1-\tau}}\frac{\partial x_\tau/\sqrt{1-\tau}}{\partial \tau} + \frac{\partial}{\partial \tau}\sqrt{\tau}\overline{s}_\tau^{\star}(\sqrt{1-\tau}y)\big|_{y=x_\tau/\sqrt{1-\tau}}.
\end{align}
Direct calculation reveals that the Jacobian matrix of $\overline{s}_\tau^{\star}(x)$ is
\begin{align}\label{eq:Jacobian}
	J_\tau(x) \coloneqq \nabla_{x}\overline{s}_\tau^{\star}(x) = -\frac{1}{\tau}I_d + \frac{1}{\tau}\Sigma_\tau(x),
\end{align}
where $\Sigma_\tau(x)$ is the conditional covariance matrix of the standard Gaussian noise $Z \sim N(0, I_d)$ conditioned on $\sqrt{1-\tau}X_0 + \sqrt{\tau}Z = x$ (cf. \eqref{eq:def-Sigma}).
Then the two terms on the right-hand side of \eqref{eq:proof-lem-bound-derivatives-temp-1} can be rewritten as follows:
\begin{subequations}
	\begin{align}
		\frac{\partial}{\partial y}\sqrt{\tau}\overline{s}_\tau^\star(\sqrt{1-\tau}y)\big|_{y=x_\tau/\sqrt{1-\tau}}\frac{\partial x_\tau/\sqrt{1-\tau}}{\partial \tau} 
		&\stackrel{\eqref{eq:ODE}~\text{and}~\eqref{eq:Jacobian}}{=}-\frac{\sqrt{\tau}}{2(1-\tau)}J_\tau(x_\tau)\overline{s}_\tau^{\star}(x_\tau)\notag\\
		&= \frac{1}{2\sqrt{\tau}(1-\tau)}\overline{s}_\tau^{\star}(x_\tau) -\frac{1}{2\sqrt{\tau}(1-\tau)}\Sigma_\tau(x_\tau)\overline{s}_\tau^{\star}(x_\tau),\label{eq:proof-lem-bound-derivatives-temp-2_part_1}\\
		\frac{\partial}{\partial \tau}\sqrt{\tau}\overline{s}_\tau^\star(\sqrt{1-\tau}y) 
		&\stackrel{\eqref{eq:score_continuous}}{=}-\frac{\partial}{\partial \tau}\sqrt{\frac{1-\tau}{\tau}}\int(y-x_0)p_{X_0|\overline{X}_\tau}(x_0|\sqrt{1-\tau}y)\mathrm{d} x_0 \notag\\
		&= - \frac{\overline{s}_\tau^{\star}(\sqrt{1-\tau}y)}{2\sqrt{\tau}(1-\tau)}  - \sqrt{\frac{1-\tau}{\tau}}\frac{\partial }{\partial \tau}\int(y-x_0)p_{X_0|\overline{X}_\tau}(x_0|\sqrt{1-\tau}y)\mathrm{d} x_0.\label{eq:proof-lem-bound-derivatives-temp-2_part_2}
	\end{align}
\end{subequations}
Combining \eqref{eq:proof-lem-bound-derivatives-temp-1}, \eqref{eq:proof-lem-bound-derivatives-temp-2_part_1} and \eqref{eq:proof-lem-bound-derivatives-temp-2_part_2}, we obtain
\begin{align*}
	\frac{\partial \sqrt{\tau}\overline{s}_{\tau}^{\star}(x_\tau)}{\partial \tau} 
	&= -\frac{1}{2\sqrt{\tau}(1-\tau)}\Sigma_\tau(x_\tau)\overline{s}_\tau^{\star}(x_\tau) - \sqrt{\frac{1-\tau}{\tau}}\frac{\partial }{\partial \tau}\int(y-x_0)p_{X_0|\overline{X}_\tau}(x_0|\sqrt{1-\tau}y)\mathrm{d} x_0.
\end{align*}
which, together with the inequality $\|a + b\|_2^2 \leq 2\|a\|_2^2 + 2\|b\|_2^2$, further leads to
\begin{align}\label{eq509}
	\mathbb{E}_{x_\tau\sim\overline{X}_\tau}\left[\left\|\frac{\partial \sqrt{\tau}\overline{s}_{\tau}^{\star}(x_\tau)}{\partial \tau}\right\|_2^2\right]
	&\le \frac{1}{2\tau(1-\tau)^2}\mathbb{E}_{x_\tau\sim\overline{X}_\tau}\left[\left\|\Sigma_\tau(x_\tau)\overline{s}_\tau^{\star}(x_\tau)\right\|_2^2\right] \notag\\
	&\quad + \frac{2(1-\tau)}{\tau}\mathbb{E}_{\sqrt{1-\tau}y\sim\overline{X}_\tau}\left[\left\|\frac{\partial }{\partial \tau}\int(y-x_0)p_{X_0|\overline{X}_\tau}(x_0|\sqrt{1-\tau}y)\mathrm{d} x_0\right\|_2^2\right].
\end{align}
In the following, we proceed to control these two terms separately.
\begin{itemize}
	\item \textbf{Bounding $\mathbb{E}_{x_\tau\sim\overline{X}_\tau}\left[\left\|\Sigma_\tau(x_\tau)\overline{s}_\tau^{\star}(x_\tau)\right\|_2^2\right]$.}
	We define the following set:
	\begin{align}\label{eq:def-set-S}
		\mathcal{S}_\tau: =
		\{x: -\log p_{\overline{X}_\tau}(x)\le \theta d\log T\},
	\end{align}
	where $\theta$ is a constant satisfying $\theta \geq c_R + 17$.
	Then we can decompose  $\mathbb{E}_{x_\tau\sim\overline{X}_\tau}[\left\|\Sigma_\tau(x_\tau)\overline{s}_\tau^{\star}(x_\tau)\right\|_2^2]$ into two terms:
	\begin{align}\label{eq:proof-lem-bound-derivative-temp-6}
		\mathbb{E}_{x_\tau\sim\overline{X}_\tau}\left[\left\|\Sigma_\tau(x_\tau)\overline{s}_\tau^{\star}(x_\tau)\right\|_2^2\right] = \int_{\mathcal{S}_\tau} \left\|\Sigma_\tau(x)\overline{s}_\tau^{\star}(x)\right\|_2^2 p_{\overline{X}_\tau}(x) \mathrm{d} x + \int_{\mathcal{S}_\tau^{\rm c}} \left\|\Sigma_\tau(x)\overline{s}_\tau^{\star}(x)\right\|_2^2 p_{\overline{X}_\tau}(x) \mathrm{d} x.
	\end{align}
	The following lemma allows us to bound $\left\|\Sigma_\tau(x)\overline{s}_\tau^{\star}(x)\right\|_2$ for $x\in\mathcal{S}_{\tau}$ and the expectation $\mathbb{E}_{x\sim \overline{X}_\tau}\left[\|\Sigma_\tau(x)\overline{s}_\tau^{\star}(x)\|_2^4\right]$. The proof is deferred to Section \ref{sec:proof-lem:technical_1}.
	\begin{lemma}\label{lem:technical_1}
		The following inequalities hold:
		\begin{subequations}\label{eq:bound-Sigma-s}
			\begin{align}
				&\left\|\overline{s}_\tau^{\star}(x)\right\|_2^2\le \frac{25(\theta+c_0)d\log T}{\tau}\quad \mathrm{and}\quad \|\Sigma_\tau(x)\|_{\mathsf{op}}\le 12(\theta+c_0)d\log T,\qquad \forall x\in\mathcal{S}_\tau,\label{eq:bound-Sigma-s-Stau}\\
				&\mathbb{E}_{x\sim \overline{X}_\tau}\left[\|\Sigma_\tau(x)\overline{s}_\tau^{\star}(x)\|_2^4\right]\le\frac{d^6}{\tau^2}.\label{eq:bound-Sigma-s-Stauc}
			\end{align}
		\end{subequations}
		Here, $c_0$ is defined in \eqref{eq:learning-rate}.
	\end{lemma}
	By virtue of \eqref{eq:bound-Sigma-s-Stau}, the integral over the set $\mathcal{S}_\tau$ can be bounded by 
	\begin{align}\label{eq:proof-lem-bound-derivative-temp-7}
		\int_{\mathcal{S}_\tau} \left\|\Sigma_\tau(x)\overline{s}_\tau^{\star}(x)\right\|_2^2 p_{\overline{X}_\tau}(x) \mathrm{d} x\lesssim \frac{(\theta+c_0)d\log T}{\tau}\int \left\|\Sigma_\tau(x)\right\|_{\mathsf{op}}^2 p_{\overline{X}_\tau}(x) \mathrm{d} x.
	\end{align}
	For the integral over $\mathcal{S}_{\tau}^{\rm c}$, one can apply Cauchy-Schwarz inequality to obtain
	\begin{align}\label{eq:proof-lem-bound-derivative-temp-5}
		\int_{\mathcal{S}_\tau^{\rm c}} \left\|\Sigma_\tau(x)\overline{s}_\tau^{\star}(x)\right\|_2^2 p_{\overline{x}_\tau}(x) \mathrm{d} x\le\sqrt{\mathbb{E}_{x\sim \overline{X}_\tau}\left[\|\Sigma_\tau(x)\overline{s}_\tau^{\star}(x)\|_2^4\right] \mathbb{P}\left\{\overline{X}_\tau\in\mathcal{S}_\tau^{\rm c}\right\}}.
	\end{align}
	To control the right-hand side of \eqref{eq:proof-lem-bound-derivative-temp-5}, we prove that $\mathcal{S}_\tau$ is a high-probability set, meaning that the probability $\mathbb{P}\left\{\overline{X}_\tau\in\mathcal{S}_\tau^{\rm c}\right\}$ is small. To this end, we define $\mathcal{B}=\{x:\|x\|_2\le \sqrt{1-\tau}T^{c_R+8} + 10\sqrt{\tau d\log T}\}$. Then we have 
	\begin{align}\label{eq:proof-prob-setEc}
		\mathbb{P}\left\{\overline{X}_\tau\in\mathcal{S}_\tau^{\rm c}\right\} &\le \int_{\mathcal{S}_\tau^{\rm c}\cap\mathcal{B}} p_{\overline{X}_\tau}(x)\mathrm{d}x + \int_{\mathcal{B}^{\rm c}} p_{\overline{X}_\tau}(x)\mathrm{d}x \notag\\
		&\le \left(2\sqrt{1-\tau}T^{c_R+8} + 20\sqrt{\tau d\log T}\right)^d\exp(-\theta d\log T) + \mathbb{P}\left\{\|X_0\|_2\ge T^{c_R+8}\right\}\notag\\
		&\quad + \mathbb{P}\left\{\|\sqrt{\tau}W\|_2\ge 10\sqrt{\tau d\log T}\right\}\notag\\
		&\le \exp(-(\theta-c_R-9)d\log T) + \frac{\mathbb{E}[\|X_0\|_2]}{T^{c_R+8}} + 2\exp\left(-8\log T\right)\lesssim \frac{1}{T^8},
	\end{align}
	provided that $T\ge 20\sqrt{d\log T}$ and $\theta\ge c_R + 17$.
	In addition, 
	Eqn.~\eqref{eq:proof-lem-bound-derivative-temp-5} together with \eqref{eq:bound-Sigma-s-Stauc} and \eqref{eq:proof-prob-setEc} yields
	\begin{align}\label{eq:proof-lem-bound-derivative-temp-8}
		\int_{\mathcal{S}_\tau^{\rm c}} \left\|\Sigma_\tau(x)\overline{s}_\tau^{\star}(x)\right\|_2^2 p_{\overline{X}_\tau}(x) \mathrm{d} x\lesssim \frac{d^3}{\tau T^4}\le \frac{d}{\tau},
	\end{align}
	as long as $T\ge \sqrt{d}$.
	Putting \eqref{eq:proof-lem-bound-derivative-temp-6}, \eqref{eq:proof-lem-bound-derivative-temp-7} and \eqref{eq:proof-lem-bound-derivative-temp-8} together, we obtain the desired bound
	\begin{align}\label{eq510}
		\mathbb{E}_{x_\tau\sim\overline{X}_\tau}\left[\left\|\Sigma_\tau(x_\tau)\overline{s}_\tau^{\star}(x_\tau)\right\|_2^2\right]
		\lesssim \frac{(\theta+c_0)d\log T}{\tau}\int \left\|\Sigma_\tau(x)\right\|_{\mathsf{op}}^2 p_{\overline{X}_\tau}(x) \mathrm{d} x + \frac{d}{\tau}.
	\end{align}
	\item \textbf{Controlling $\mathbb{E}_{\sqrt{1-\tau}y\sim\overline{X}_\tau}\left[\left\|\frac{\partial }{\partial \tau}\int(y-x_0)p_{X_0|\overline{X}_\tau}(x_0|\sqrt{1-\tau}y)\mathrm{d} x_0\right\|_2^2\right]$.}
	We define
	\begin{align}\label{def:tilde_X_t}
		t \coloneqq \frac{\tau}{1-\tau}~\quad~\text{and}~\quad~\widetilde{X}_t = X_0 + \sqrt{t} Z~\quad~\text{where}~\quad~Z \sim\mathcal{N}(0,I_d).
	\end{align}
	Then it is straightforward to verify that
	\begin{align*}
		\frac{\partial}{\partial t}p_{\widetilde{X}_t|X_0}(y|x_0) = \frac{\|y - x_0\|_2^2}{2t^2}p_{\widetilde{X}_t|X_0}(y|x_0)
	\end{align*}
	and
	\begin{align*}
		\frac{\partial}{\partial t}\int p_{\widetilde{X}_t|X_0}(y|x_0)p_{X_0}(x_0) \mathrm{d} x_0 = \int \left(\frac{\partial}{\partial t}p_{\widetilde{X}_t|X_0}(y|x_0)\right)p_{X_0}(x_0) \mathrm{d} x_0 = \int\frac{\|y - x_0\|_2^2}{2t^2} p_{\widetilde{X}_t|X_0}(y|x_0)p_{X_0}(x_0) \mathrm{d} x_0.
	\end{align*}
	Therefore, one can write
	\begin{align}\label{eq501}
		&\frac{\partial}{\partial \tau}\int_{x_0}(y-x_0)p_{X_0|\overline{X}_\tau}(x_0|\sqrt{1-\tau}y)\mathrm d x_0 \notag\\
		&\quad=\frac{\partial}{\partial \tau}t\cdot \frac{\partial}{\partial t} \int(y-x_0) p_{X_0|\widetilde{X}_t}(x_0|y) \mathrm{d} x_0 \notag\\
		&\quad=-\frac{1}{(1-\tau)^2}\frac{\partial}{\partial t} \frac{\int(y-x_0)p_{\widetilde{X}_t|X_0}(y|x_0)p_{X_0}(x_0) \mathrm{d} x_0}{\int p_{\widetilde{X}_t|X_0}(y|x_0)p_{X_0}(x_0) \mathrm{d} x_0} \notag\\
		&\quad=-\frac{1}{(1-\tau)^2} \Bigg(\frac{\frac{\partial}{\partial t}\int(y-x_0)p_{\widetilde{X}_t|X_0}(y|x_0)p_{X_0}(x_0) \mathrm{d} x_0}{\int p_{\widetilde{X}_t|X_0}(y|x_0)p_{X_0}(x_0) \mathrm{d} x_0}\notag\\ &\hspace{3cm}- \frac{\int(y-x_0)p_{\widetilde{X}_t|X_0}(y|x_0)p_{X_0}(x_0) \mathrm{d} x_0\frac{\partial}{\partial t}\int p_{\widetilde{X}_t|X_0}(y|x_0)p_{X_0}(x_0) \mathrm{d} x_0}{\left(\int p_{\widetilde{X}_t|X_0}(y|x_0)p_{X_0}(x_0) \mathrm{d} x_0\right)^2}\Bigg)\notag\\
		&\quad= -\frac{1}{(1-\tau)^2}\left(\int \frac{\|y-x_0\|_2^2}{2t^2}(y-x_0) p_{X_0|\widetilde{X}_t}(x_0|y)\mathrm{d}x_0
		 - \int (y-x_0)p_{X_0|\widetilde{X}_t}(x_0|y)\mathrm{d}x_0 \int \frac{\|y-x_0\|_2^2}{2t^2} p_{X_0|\widetilde{X}_t}(x_0|y)\mathrm{d}x_0\right).
	\end{align}
	Recognizing that $p_{X_0|\widetilde{X}_t}(x_0|y)\mathrm{d}x_0 = p_{Z|X_0+\sqrt{t}Z}(\frac{y-x_0}{\sqrt{t}}|y)\mathrm{d}(\frac{y-x_0}{\sqrt{t}})$, we have
	\begin{align}\label{eq502}
		\int \frac{\|y-x_0\|_2^2}{2t^2}(y-x_0) p_{X_0|\widetilde{X}_t}(x_0|y)\mathrm{d}x_0 \stackrel{z = \frac{y-x_0}{\sqrt{t}}}{=} \int \frac{\|z\|_2^2}{2t}\sqrt{t}z p_{Z|X_0+\sqrt{t}Z}(z|y)\mathrm{d}z = \frac{1}{2\sqrt{t}}\mathbb{E}\big[\|Z\|_2^2Z|X_0 + \sqrt{t}Z = y\big].
	\end{align}
	Similarly, one has
	\begin{align}\label{eq503}
		&\int (y-x_0)p_{X_0|\widetilde{X}_t}(x_0|y)\mathrm{d}x_0 \int \frac{\|y-x_0\|_2^2}{2t^2} p_{X_0|\widetilde{X}_t}(x_0|y)\mathrm{d}x_0\notag\\ &\quad= \sqrt{t}\mathbb{E}\big[Z|X_0 + \sqrt{t}Z = y\big]\cdot\frac{1}{2t}\mathbb{E}\big[\|Z\|_2^2|X_0 + \sqrt{t}Z = y\big]\notag\\ &\quad= \frac{1}{2\sqrt{t}}\mathbb{E}\big[Z|X_0 + \sqrt{t}Z = y\big]\mathbb{E}\big[\|Z\|_2^2|X_0 + \sqrt{t}Z = y\big].
	\end{align}
	For notational convenience, we write
	\begin{align*}
		\mathbb{E}_{Z|y}[\cdot] := \mathbb{E}[\cdot|X_0 + \sqrt{t}Z = y],
	\end{align*}
	i.e., the expectation conditioned on $X_0 + \sqrt{t}Z = y$.
	Combining \eqref{eq501} - \eqref{eq503} and \eqref{def:tilde_X_t} yields
	\begin{align*}
		&\frac{\partial}{\partial \tau}\int_{x_0}(y-x_0)p_{X_0|\overline{X}_\tau}(x_0|\sqrt{1-\tau}y)\mathrm d x_0 \notag\\
		&\quad=-\frac{1}{2\tau^{1/2}(1-\tau)^{3/2}}\mathbb{E}_{Z|y}\left[\|Z\|_2^2Z-\mathbb{E}_{Z|y}[Z]\mathbb{E}_{Z|y}[\|Z\|_2^2]\right]  \notag\\
		&\quad= -\frac{1}{2\tau^{1/2}(1-\tau)^{3/2}}\mathbb{E}_{Z|y}\left[\left(Z-\mathbb{E}_{Z|y}[Z]\right)\left(\|Z\|_2^2 - \mathbb{E}_{Z|y}[\|Z\|_2^2]\right)\right].
	\end{align*}
	Consequently, one has
	\begin{align}\label{eq506}
		&\mathbb{E}_{\sqrt{1-\tau}y\sim\overline{X}_\tau}\left[\left\|\frac{\partial }{\partial \tau}\int(y-x_0)p_{X_0|\overline{X}_\tau}(x_0|\sqrt{1-\tau}y)\mathrm{d} x_0\right\|_2^2\right]\notag\\
		&\quad = \frac{1}{\tau(1-\tau)^3}\mathbb{E}_{\sqrt{1-\tau}y\sim\overline{X}_\tau}\left[\left\|\mathbb{E}_{Z|y}\left[\left(Z-\mathbb{E}_{Z|y}[Z]\right)\left(\|Z\|_2^2 - \mathbb{E}_{Z|y}[\|Z\|_2^2]\right)\right]\right\|_2^2\right].
	\end{align}
	We claim for the moment that 
	\begin{align}\label{eq507}
		\mathbb{E}_{\sqrt{1-\tau}y\sim\overline{X}_\tau}\left[\left\|\mathbb{E}_{Z|y}\left[\left(Z-\mathbb{E}_{Z|y}[Z]\right)\left(\|Z\|_2^2 - \mathbb{E}_{Z|y}[\|Z\|_2^2]\right)\right]\right\|_2^2\right] \lesssim d\min\left\{d\log T,L\right\};
	\end{align}
    we defer the proof to end of the section. Combining \eqref{eq506} and \eqref{eq507}, we have
    \begin{align}\label{eq508}
    	\mathbb{E}_{\sqrt{1-\tau}y\sim\overline{X}_\tau}\left[\left\|\frac{\partial }{\partial \tau}\int(y-x_0)p_{X_0|\overline{X}_\tau}(x_0|\sqrt{1-\tau}y)\mathrm{d} x_0\right\|_2^2\right] \leq \frac{d\min\left\{d\log T,L\right\}}{\tau(1-\tau)^3}.
    \end{align}
\end{itemize}

Putting \eqref{eq509}, \eqref{eq510} and \eqref{eq508} together, we arrive at
\begin{align}
	\mathbb{E}_{x_\tau\sim\overline{X}_\tau}\left[\left\|\frac{\partial \sqrt{\tau}\overline{s}_{\tau}^{\star}(x_\tau)}{\partial \tau}\right\|_2^2\right]&\lesssim \frac{(\theta+c_0)d}{\tau^2(1-\tau)^2}\left(\log T\int \left\|\Sigma_\tau(x)\right\|_2^2 p_{\overline{X}_\tau}(x) \mathrm{d} x + 1\right) +\frac{d\min\left\{d\log T,L\right\}}{\tau^2(1-\tau)^2}.
\end{align}
\paragraph{Proof of Claim \eqref{eq507}.} It remains to validate \eqref{eq507}. By virtue of Cauchy-Schwarz inequality, we have
\begin{align}\label{eq504}
	&\left\|\mathbb{E}_{Z|y}\left[\left(Z-\mathbb{E}_{Z|y}[Z]\right)\left(\|Z\|_2^2 - \mathbb{E}_{Z|y}[\|Z\|_2^2]\right)\right]\right\|_2^2\notag\\
	&\quad = \sup_{u \in \mathbb{S}^{d-1}}\left(u^\top\mathbb{E}_{Z|y}\left[\left(Z-\mathbb{E}_{Z|y}[Z]\right)\left(\|Z\|_2^2 - \mathbb{E}_{Z|y}[\|Z\|_2^2]\right)\right]\right)^2\notag\\
	&\quad\le \sup_{u \in \mathbb{S}^{d-1}}\mathbb{E}_{Z|y}\left[\left(u^\top\left(Z-\mathbb{E}_{Z|y}[Z]\right)\right)^2\right]\mathbb{E}_{Z|y}\left[\left(\|Z\|_2^2 - \mathbb{E}_{Z|y}[\|Z\|_2^2]\right)^2\right]\notag\\
	&\quad = \sup_{u \in \mathbb{S}^{d-1}}u^\top\mathbb{E}_{Z|y}\left[\left(Z-\mathbb{E}_{Z|y}[Z]\right)\left(Z-\mathbb{E}_{Z|y}[Z]\right)^\top\right]u\cdot\left(\mathbb{E}_{Z|y}\|Z\|_2^4 - \left(\mathbb{E}_{Z|y}[\|Z\|_2^2]\right)^2\right)\notag\\
	&\quad=\sup_{u \in \mathbb{S}^{d-1}}u^\top\mathsf{Cov}\left(Z|X_0 + \sqrt{t}Z = y\right)u\cdot\left(\mathbb{E}_{Z|y}\|Z\|_2^4 - \left(\mathbb{E}_{Z|y}[\|Z\|_2^2]\right)^2\right)\notag\\
	&\quad\stackrel{\eqref{eq:def-Sigma}}{=}\sup_{u \in \mathbb{S}^{d-1}}u^\top\left(\Sigma_\tau(\sqrt{1-\tau}y)\right)u\cdot\left(\mathbb{E}_{Z|y}\|Z\|_2^4 - \left(\mathbb{E}_{Z|y}[\|Z\|_2^2]\right)^2\right)\notag\\
	&\quad\leq \left\|\Sigma_\tau(\sqrt{1-\tau}y)\right\|_2\left(\mathbb{E}_{Z|y}\|Z\|_2^4 - \left(\mathbb{E}_{Z|y}[\|Z\|_2^2]\right)^2\right),
\end{align}
where $\mathbb{S}^{d-1} = \{x|x \in \mathbb{R}^d, \|x\|_2 = 1\}$. We consider two scenarios: $L<d\log T$ and $L\ge d\log T$.
\paragraph{Case 1: $L<d\log T$.} In view of \eqref{eq504}, we have
\begin{align}\label{eq505}
	&\mathbb{E}_{\sqrt{1-\tau}y\sim \overline{X}_\tau}\left[\left\|\mathbb{E}_{Z|y}\left[\left(Z-\mathbb{E}_{Z|y}[Z]\right)\left(\|Z\|_2^2 - \mathbb{E}_{Z|y}[\|Z\|_2^2]\right)\right]\right\|_2^2\ind(\|\Sigma_\tau(\sqrt{1-\tau}y)\|_2\le L+1)\right] \notag\\
	&\quad\leq \mathbb{E}_{\sqrt{1-\tau}y\sim \overline{X}_\tau}\left[\left\|\Sigma_\tau(x_\tau)\right\|_2\left(\mathbb{E}_{Z|y}\|Z\|_2^4 - \left(\mathbb{E}_{Z|y}[\|Z\|_2^2]\right)^2\right)\ind(\|\Sigma_\tau(\sqrt{1-\tau}y)\|_2\le L+1)\right]\notag\\
	&\quad\leq (L+1)\mathbb{E}_{\sqrt{1-\tau}y\sim \overline{X}_\tau}\left[ \mathbb{E}_{Z|y}\|Z\|_2^4 - \left(\mathbb{E}_{Z|y}[\|Z\|_2^2]\right)^2\right]\notag\\
	&\quad = (L+1)\left(\mathbb{E}[\|Z\|_2^4] - \mathbb{E}_{\sqrt{1-\tau}y\sim \overline{X}_\tau}\left[\left(\mathbb{E}_{Z|y}[\|Z\|_2^2]\right)^2\right]\right)\notag\\
	&\stackrel{\text{Jensen's inequality}}{\leq} (L+1)\left(\mathbb{E}[\|Z\|_2^4] - \left(\mathbb{E}[\|Z\|_2^2]\right)^2\right)\notag\\
	&\quad= (L+1)(d^2 + 2d - d^2) = 2d(L+1).
\end{align}
Furthermore, using Cauchy-Schwarz inequality, one can show that
\begin{align}\label{eq511}
	&\mathbb{E}_{\sqrt{1-\tau}y\sim \overline{X}_\tau}\left[\left\|\mathbb{E}_{Z|y}\left[\left(Z-\mathbb{E}_{Z|y}[Z]\right)\left(\|Z\|_2^2 - \mathbb{E}_{Z|y}[\|Z\|_2^2]\right)\right]\right\|_2^2\ind(\|\Sigma_\tau(\sqrt{1-\tau}y)\|_2 > L)\right]\notag\\
	&\quad \leq \left(\mathbb{E}_{\sqrt{1-\tau}y\sim \overline{X}_\tau}\left[\left\|\mathbb{E}_{Z|y}\left[\left(Z-\mathbb{E}_{Z|y}[Z]\right)\left(\|Z\|_2^2 - \mathbb{E}_{Z|y}[\|Z\|_2^2]\right)\right]\right\|_2^4\right]\right)^{1/2}\sqrt{\mathbb{P}_{\sqrt{1-\tau}y\sim \overline{X}_\tau}(\|\Sigma_\tau(\sqrt{1-\tau}y)\|_2>L)}\notag\\
	&\quad \leq \left(\mathbb{E}_{\sqrt{1-\tau}y\sim \overline{X}_\tau}\left[\mathbb{E}_{Z|y}\left[\|Z - \mathbb{E}_{Z|y}[Z]\|_2^8\right]\cdot\mathbb{E}_{Z|y}\left[\left(\|Z\|_2^2 - \mathbb{E}_{Z|y}[\|Z\|_2^2]\right)^8\right]\right]\right)^{1/4}\sqrt{\mathbb{P}_{x_{\tau}\sim \overline{X}_\tau}(\|\Sigma_\tau(x_{\tau})\|_2>L)}\notag\\
	&\quad \leq \left(\mathbb{E}_{\sqrt{1-\tau}y\sim \overline{X}_\tau}\left[\mathbb{E}_{Z|y}\left[\|Z - \mathbb{E}_{Z|y}[Z]\|_2^{16}\right]\right]\cdot\mathbb{E}_{\sqrt{1-\tau}y\sim \overline{X}_\tau}\left[\mathbb{E}_{Z|y}\left[\left(\|Z\|_2^2 - \mathbb{E}_{Z|y}[\|Z\|_2^2]\right)^{16}\right]\right]\right)^{1/8}\notag\\&\qquad\cdot\sqrt{\mathbb{P}_{x_{\tau}\sim \overline{X}_\tau}(\|\Sigma_\tau(x_{\tau})\|_2>L+1)}.
\end{align}
Applying Jensen's inequality yields
\begin{align}\label{eq512}
	&\mathbb{E}_{\sqrt{1-\tau}y\sim \overline{X}_\tau}\left[\mathbb{E}_{Z|y}\left[\|Z - \mathbb{E}_{Z|y}[Z]\|_2^{16}\right]\right]\notag\\
	&\quad \lesssim \mathbb{E}_{\sqrt{1-\tau}y\sim \overline{X}_\tau}\left[\mathbb{E}_{Z|y}\left[\|Z\|_2^{16}\right] + \mathbb{E}_{Z|y}\left[\|\mathbb{E}_{Z|y}[Z]\|_2^{16}\right]\right]\notag\\
	& \quad \leq \mathbb{E}_{\sqrt{1-\tau}y\sim \overline{X}_\tau}\left[\mathbb{E}_{Z|y}\left[\|Z\|_2^{16}\right] + \mathbb{E}_{Z|y}\left[\mathbb{E}_{Z|y}\left[\|Z\|_2^{16}\right]\right]\right]\notag\\
	&\quad= 2\mathbb{E}\left[\|Z\|_2^{16}\right] \lesssim d^8.
\end{align}
Similarly, one has
\begin{align}\label{eq513}
	\mathbb{E}_{\sqrt{1-\tau}y\sim \overline{X}_\tau}\left[\mathbb{E}_{Z|y}\left[\left(\|Z\|_2^2 - \mathbb{E}_{Z|y}[\|Z\|_2^2]\right)^{16}\right]\right] \lesssim \mathbb{E}\left[\|Z\|_{2}^{32}\right] \leq d^{16}.
\end{align}
By virtue of \cite[Eqn.~(25b)]{li2024sharp}, we have
\begin{align*}
	\Sigma_\tau(x_{\tau}) = \tau\nabla \overline{s}_{\tau}^\star(x_{\tau}) + I_d,
\end{align*}
which, together with Definition~\ref{def:score-lipschitz}, implies
\begin{align}\label{eq514}
	\mathbb{P}_{x_{\tau}\sim \overline{X}_\tau}(\|\Sigma_\tau(x_{\tau})\|_{\mathsf{op}}>L+1) \leq \mathbb{P}_{x_{\tau}\sim \overline{X}_\tau}\left(\left\|\tau\nabla \overline{s}_{\tau}^\star(x_{\tau})\right\|_{\mathsf{op}}>L\right) \leq d^{-4}.
\end{align}
Combining \eqref{eq511} - \eqref{eq514}, one has
\begin{align}\label{eq515}
	\mathbb{E}_{\sqrt{1-\tau}y\sim \overline{X}_\tau}\left[\left\|\mathbb{E}_{Z|y}\left[\left(Z-\mathbb{E}_{Z|y}[Z]\right)\left(\|Z\|_2^2 - \mathbb{E}_{Z|y}[\|Z\|_2^2]\right)\right]\right\|_2^2\ind(\|\Sigma_\tau(\sqrt{1-\tau}y)\|_{\mathsf{op}} > L)\right] \lesssim d.
\end{align}
Eqn.~\eqref{eq505} and \eqref{eq515} together show that
\begin{align}\label{eq519}
	&\quad \mathbb{E}_{\sqrt{1-\tau}y\sim \overline{X}_\tau}\left[\left\|\mathbb{E}_{Z|y}\left[\left(Z-\mathbb{E}_{Z|y}[Z]\right)\left(\|Z\|_2^2 - \mathbb{E}_{Z|y}[\|Z\|_2^2]\right)\right]\right\|_2^2\right]\lesssim dL.
\end{align}

\paragraph{Case 2: $L\ge d\log T$.} We set $\theta = c_R + 17$, where $c_R$ is defined in \eqref{assump:bounded_second_moment}. Recalling the definition of $\mathcal{S}_\tau$ in \eqref{eq:def-set-S} and applying Lemma~\ref{lem:technical_1} and \eqref{eq504}, one has
\begin{align}\label{eq516}
	&\mathbb{E}_{\sqrt{1-\tau}y\sim \overline{X}_\tau}\left[\left\|\mathbb{E}_{Z|y}\left[\left(Z-\mathbb{E}_{Z|y}[Z]\right)\left(\|Z\|_2^2 - \mathbb{E}_{Z|y}[\|Z\|_2^2]\right)\right]\right\|_2^2\ind(\sqrt{1-\tau}y\in\mathcal{S}_\tau)\right] \notag\\
	&\quad \leq \mathbb{E}_{\sqrt{1-\tau}y\sim \overline{X}_\tau}\left[\left\|\Sigma_\tau(\sqrt{1-\tau}y)\right\|_{\mathsf{op}}\left(\mathbb{E}_{Z|y}\|Z\|_2^4 - \left(\mathbb{E}_{Z|y}[\|Z\|_2^2]\right)^2\right)\ind(\sqrt{1-\tau}y\in\mathcal{S}_\tau)\right] \notag\\ 
	&\quad \leq 12(\theta + c_0)d\log T\mathbb{E}_{\sqrt{1-\tau}y\sim \overline{X}_\tau}\left[\left(\mathbb{E}_{Z|y}\|Z\|_2^4 - \left(\mathbb{E}_{Z|y}[\|Z\|_2^2]\right)^2\right)\right]\notag\\
	&\quad = 12(\theta + c_0)d\log T\left(\mathbb{E}\left[\|Z\|_2^4\right] - \mathbb{E}_{\sqrt{1-\tau}y\sim \overline{X}_\tau}\left[\left(\mathbb{E}_{Z|y}[\|Z\|_2^2]\right)^2\right]\right)\notag\\
	&\stackrel{\text{Jensen's inequality}}{\leq} 12(\theta + c_0)d\log T\left(\mathbb{E}[\|Z\|_2^4] - \left(\mathbb{E}[\|Z\|_2^2]\right)^2\right)\notag\\
	&\quad\leq 12(\theta + c_0)d\log T(d^2 + 2d - d^2)\notag\\
	&\quad\lesssim d^2\log T.
\end{align}
Repeating similar arguments as in \eqref{eq515} yields 
\begin{align*}
	&\mathbb{E}_{\sqrt{1-\tau}y\sim \overline{X}_\tau}\left[\left\|\mathbb{E}_{Z|y}\left[\left(Z-\mathbb{E}_{Z|y}[Z]\right)\left(\|Z\|_2^2 - \mathbb{E}_{Z|y}[\|Z\|_2^2]\right)\right]\right\|_2^2\ind(\sqrt{1-\tau}y\in\mathcal{S}_\tau^{\rm c})\right]\\ &\quad\leq \left(\mathbb{E}_{\sqrt{1-\tau}y\sim \overline{X}_\tau}\left[\mathbb{E}_{Z|y}\left[\|Z - \mathbb{E}_{Z|y}[Z]\|_2^{16}\right]\right]\cdot\mathbb{E}_{\sqrt{1-\tau}y\sim \overline{X}_\tau}\left[\mathbb{E}_{Z|y}\left[\left(\|Z\|_2^2 - \mathbb{E}_{Z|y}[\|Z\|_2^2]\right)^{16}\right]\right]\right)^{1/8}\cdot\sqrt{\mathbb{P}_{x_{\tau}\sim \overline{X}_\tau}\left(x_{\tau} \in \mathcal{S}_\tau^{\rm c}\right)}\notag\\
	&\quad\lesssim d^3\sqrt{\mathbb{P}_{x_{\tau}\sim \overline{X}_\tau}\left(x_{\tau} \in \mathcal{S}_\tau^{\rm c}\right)}.
\end{align*}
By virtue of \eqref{eq:proof-prob-setEc}, we have
\begin{align}\label{eq517}
	&\mathbb{E}_{\sqrt{1-\tau}y\sim \overline{X}_\tau}\left[\left\|\mathbb{E}_{Z|y}\left[\left(Z-\mathbb{E}_{Z|y}[Z]\right)\left(\|Z\|_2^2 - \mathbb{E}_{Z|y}[\|Z\|_2^2]\right)\right]\right\|_2^2\ind(\sqrt{1-\tau}y\in\mathcal{S}_\tau^{\rm c})\right]\notag\\
	&\quad \lesssim d^3\cdot T^{-4} \leq d,
\end{align}
provided that $T \geq \sqrt{d}$.
Putting \eqref{eq516} and \eqref{eq517} together leads to
\begin{align}\label{eq518}
	\mathbb{E}_{\sqrt{1-\tau}y\sim \overline{X}_\tau}\left[\left\|\mathbb{E}_{Z|y}\left[\left(Z-\mathbb{E}_{Z|y}[Z]\right)\left(\|Z\|_2^2 - \mathbb{E}_{Z|y}[\|Z\|_2^2]\right)\right]\right\|_2^2\ind(\sqrt{1-\tau}y\in\mathcal{S}_\tau)\right] \lesssim d^2\log T.
\end{align}
Combining \eqref{eq519} and \eqref{eq518} finishes the proof of Claim \eqref{eq507}.

\section{Proof of Lemma \ref{lem:endpoints}}
\label{sec:proof-lem-endpoints}
The first inequality comes from Lemma 3 in \citet{li2024sharp}.
It suffices to prove the second one.
To this end, we consider two scenarios: $d>L^2$ and $d\le L^2$.
\begin{itemize}
	\item \emph{Case 1: $d>L^2$.}
	Let $\mathcal{L}_\tau$ denote the set of $x$ such that the operator norm of the covariance matrix $\Sigma_\tau(x)$ is at most $L$, i.e.,
	$$
	\mathcal{L}_\tau\coloneqq\{x:\|\Sigma_\tau(x)\|_{\mathsf{op}}\le L+1\}.
	$$
	By virtue of \eqref{eq514}, we know that $\mathbb{P}(\overline{X}_\tau\in\mathcal{L}_\tau^{\rm c})\le d^{-4}$, and consequently
	\begin{align*}
		\mathbb{E}_{x_\tau\sim \overline{X}_\tau} \left\|\Sigma_\tau(x_\tau)\right\|_{\mathsf{op}}^2 
		&\le (L+1)^2 + \int_{\mathcal{L}_\tau^{\rm c}} \left\|\Sigma_\tau(x)\right\|_{\mathsf{op}}^2 p_{\overline{X}_\tau}(x)\mathrm{d} x
		\overset{\text{(i)}}{\le} (L+1)^2 + \sqrt{\mathbb{E}_{x\sim \overline{X}_\tau}\left[ \left\|\Sigma_\tau(x)\right\|_{\mathsf{op}}^4\right]} \sqrt{\mathbb{P}(\overline{X}_\tau\in\mathcal{L}_\tau^{\rm c})} \notag\\
		&\overset{\text{(ii)}}{\lesssim} L^2+ \sqrt{\frac{d^4}{d^4}} \lesssim L^2,
	\end{align*}
	where (i) holds due to the Cauchy-Schwarz inequality, and (ii) makes use of the fact that 
	$$
	\mathbb{E}_{x\sim \overline{X}_\tau}\left[ \left\|\Sigma_\tau(x)\right\|_{\mathsf{op}}^4\right]\le \mathbb{E}_{x\sim \overline{X}_\tau}\left[\mathbb{E}\left[\|Z\|_2^8|\sqrt{1-\tau}X_0 + \sqrt{\tau}Z = x\right]\right] = \mathbb{E}_{Z\sim\mathcal{N}(0,I_d)}\left[\|Z\|_2^8\right]\lesssim d^4.
	$$
	Thus we have
	\begin{align}\label{ineq:integral_case_1}
		\sum_{t=2}^T\frac{\overline{\alpha}_{t}}{(1-\overline{\alpha}_{t})}\int_{\widetilde{\tau}_{T-t+2}}^{\tau_{T-t+1}}\frac{\mathbb{E}_{x_\tau\sim \overline{X}_\tau} \left\|\Sigma_\tau(x_\tau)\right\|_{\mathsf{op}}^2}{(1-\tau)^2}\mathrm{d} \tau
		&\lesssim \sum_{t=2}^T\frac{L^2(\tau_{T-t+1}-\widetilde{\tau}_{T-t+2})}{\overline{\alpha}_{t}(1-\overline{\alpha}_{t})}\notag\\
		&\overset{\eqref{eq:lem-tildetau-1}}{\lesssim} \sum_{t=1}^T\frac{L^2\overline{\alpha}_{t-1}\log T}{T\overline{\alpha}_{t}}\notag\\ &\overset{\eqref{eq:learning-rate}}{\lesssim} L^2\log T.
	\end{align}
	\item \emph{Case 2: $d\le L^2$.}
	Now we move on to the second case. By virtue of \citet[Claim (90)]{li2024sharp}, one has, for any $0<\tau_1<\tau_2<1$,
	\begin{align}\label{eq:integral_fact}
		\int_{\tau_1}^{\tau_2}\frac{\mathbb{E}[\mathsf{Tr}(\Sigma_\tau^2(x_\tau))]}{(1-\tau)^2}\mathrm{d}\tau
		&=\frac{\tau_1
		}{1-\tau_1}\mathbb{E}[\mathsf{Tr}(\Sigma_{\tau_1}(x_{\tau_1}))] - \frac{\tau_2
		}{1-\tau_2}\mathbb{E}[\mathsf{Tr}(\Sigma_{\tau_2}(x_{\tau_2}))].
	\end{align}
    We make the observation that
    \begin{align*}
    \tau_{T-t+2} - \widetilde{\tau}_{T-t+2} = \widetilde{\alpha}_{t-1} - \overline{\alpha}_{t-1}=\frac{(1-\alpha_t)(1-\overline{\alpha}_{t-1})\overline{\alpha}_{t-1}}{\overline{\alpha}_{t-1}(1-\alpha_t) + \alpha_t(1-\overline{\alpha}_{t-1})} \le\frac{\overline{\alpha}_{t-1} - \overline{\alpha}_{t}}{\alpha_t} \le 2(\tau_{T-t+1} - \tau_{T-t+2}).
    \end{align*}
    In view of \citet[Lemma 2 Part (a)]{li2024sharp} and the fact that
    $$|\tau_{T-t+1} - \widetilde{\tau}_{T-t+2}|/\tau_{T-t+1}(1-\tau_{T-t+1})\lesssim \log T/T \lesssim 1/(d\log T),$$ we have
    \begin{align*}
    \mathbb{E}_{x_\tau\sim \overline{X}_\tau} \mathsf{Tr}\left(\Sigma^2_\tau(x_\tau)\right) \lesssim \mathbb{E}_{x_\tau\sim X_{\tau_{T-t+1}}} \mathsf{Tr}\left(\Sigma^2_{\tau_{T-t+1}}(x_\tau)\right) + \frac{1}{T^{c_2d}}.
    \end{align*}
    Combining \eqref{eq:integral_fact} and the previous inequality yields
    \begin{align*}
    \int_{\widetilde{\tau}_{T-t+2}}^{\tau_{T-t+1}}\frac{\mathbb{E}_{x_\tau\sim \overline{X}_\tau} \mathsf{Tr}\left(\Sigma^2_\tau(x_\tau)\right)}{(1-\tau)^2}\mathrm{d} \tau 
    &\lesssim \frac{{\tau}_{T-t+1}-\widetilde{\tau}_{T-t+2}}{(1-\tau_{T-t+1})^2}\left(\mathbb{E}_{x_\tau\sim \overline{X}_{\tau_{T-t+1}}}\left[\mathsf{Tr}\left(\Sigma^2_{\tau_{T-t+1}}(x_\tau)\right)\right] + T^{-c_2d}\right),
    \end{align*}
    which further implies
    \begin{align*}
    \frac{\overline{\alpha}_{t-1}}{1-\overline{\alpha}_{t-1}}\int_{\widetilde{\tau}_{T-t+2}}^{\tau_{T-t+1}}\frac{\mathbb{E}_{x_\tau\sim \overline{X}_\tau} \left\|\Sigma_\tau(x_\tau)\right\|_{\mathsf{op}}^2}{(1-\tau)^2}\mathrm{d} \tau &\le \frac{\overline{\alpha}_{t-1}}{1-\overline{\alpha}_{t-1}}\int_{\widetilde{\tau}_{T-t+2}}^{\tau_{T-t+1}}\frac{\mathbb{E}_{x_\tau\sim \overline{X}_\tau} \left\|\Sigma_\tau(x_\tau)\right\|_{\rm F}^2}{(1-\tau)^2}\mathrm{d} \tau\notag\\
		&=\frac{\overline{\alpha}_{t-1}}{1-\overline{\alpha}_{t-1}}\int_{\widetilde{\tau}_{T-t+2}}^{\tau_{T-t+1}}\frac{\mathbb{E}_{x_\tau\sim \overline{X}_\tau} \mathsf{Tr}\left(\Sigma^2_\tau(x_\tau)\right)}{(1-\tau)^2}\mathrm{d} \tau\notag\\
        &\lesssim \frac{\overline{\alpha}_{t-1}-\overline{\alpha}_{t}}{\overline{\alpha}_{t-1}(1-\overline{\alpha}_{t-1})}\left(\mathbb{E}_{x_\tau\sim \overline{X}_{\tau_{T-t+1}}}\left[\mathsf{Tr}\left(\Sigma^2_{\tau_{T-t+1}}(x_\tau)\right)\right] + T^{-c_2d}\right)\notag\\
        &\asymp \frac{1-{\alpha}_{t}}{1-\overline{\alpha}_{t}}\left(\mathbb{E}_{x_\tau\sim \overline{X}_{\tau_{T-t+1}}} \left[\mathsf{Tr}\left(\Sigma^2_{\tau_{T-t+1}}(x_\tau)\right)\right] + T^{-c_2d}\right),
    \end{align*}
    where the second line arises from $\mathsf{Tr}\left(\Sigma^2_\tau(x_\tau)\right) = \|\Sigma_\tau(x_\tau)\|_{\rm F}^2$.
	Summing over $t=2, \dots, T$, we obtain
    \begin{align}\label{ineq:integral_case_2}
    \sum_{t=2}^T\frac{\overline{\alpha}_{t-1}}{1-\overline{\alpha}_{t-1}}\int_{\widetilde{\tau}_{T-t+2}}^{\tau_{T-t+1}}\frac{\mathbb{E}_{x_\tau\sim \overline{X}_\tau} \left\|\Sigma_\tau(x_\tau)\right\|_{\mathsf{op}}^2}{(1-\tau)^2}\mathrm{d} \tau 
    &\lesssim \sum_{t=2}^T \frac{1-{\alpha}_{t}}{1-\overline{\alpha}_{t}}\left(\mathbb{E}_{x_\tau\sim \overline{X}_{\tau_{T-t+1}}} \left[\mathsf{Tr}\left(\Sigma^2_{\tau_{T-t+1}}(x_\tau)\right)\right] + T^{-c_2d}\right)\notag\\
    &\lesssim d\log T + T^{-c_2d}\log T,
    \end{align}
    where the second line holds due to \citet[Lemma 2]{li2024sharp} and \eqref{eq:learning-rate}.
\end{itemize}

Putting \eqref{ineq:integral_case_1} and \eqref{ineq:integral_case_2} for the above two cases together, we know that Lemma~\ref{lem:endpoints} always holds.

\section{Proof of Lemma~\ref{lem:technical_1}}\label{sec:proof-lem:technical_1}
		The inequality \eqref{eq:bound-Sigma-s-Stau} comes from Lemma 1 in \citet{li2024d}.
		The inequality \eqref{eq:bound-Sigma-s-Stauc} makes use of the fact that
		\begin{align*}
			\mathbb{E}_{x\sim \overline{X}_\tau}\left[\|\Sigma_\tau(x)\overline{s}_\tau^{\star}(x)\|_2^4\right]&\stackrel{\text{Cauchy-Schwarz}}{\le} \sqrt{\mathbb{E}_{x\sim \overline{X}_\tau}\left[\|\Sigma_\tau(x)\|_{\mathsf{op}}^8\right]\mathbb{E}_{x\sim \overline{X}_\tau}\left[\|\overline{s}_\tau^{\star}(x)\|_2^8\right]}
			\le \frac{d^6}{\tau^2},
		\end{align*}
		where the last inequality holds due to the facts
		\begin{align}
			\mathbb{E}_{x\sim \overline{X}_\tau}\left[\|\Sigma_\tau(x)\|_{\mathsf{op}}^8\right]&=\mathbb{E}_{x\sim \overline{X}_\tau}\left[\|\mathsf{Cov}(Z|\overline{X}_\tau=x)\|_{\mathsf{op}}^8\right]\le \mathbb{E}_{x\sim \overline{X}_\tau}\left[\left(\mathbb{E}[\|Z\|_2^{2}|\overline{X}_\tau=x]\right)^8\right]\notag\\
			&\stackrel{\text{Jensen's inequality}}{\le} \mathbb{E}_{x\sim \overline{X}_\tau}\left[\mathbb{E}[\|Z\|_2^{16}|\overline{X}_\tau=x]\right] = \mathbb{E}[\|Z\|_2^{16}] \asymp d^8,\label{ineq:technical_1}\\
			\mathbb{E}_{x\sim \overline{X}_\tau}\left[\|\overline{s}_\tau^{\star}(x)\|_2^8\right]& =  \mathbb{E}_{x\sim \overline{X}_\tau}\left[\left\|\mathbb{E}\left[\frac{Z}{\sqrt{\tau}}|\overline{X}_\tau = x\right]\right\|_2^8\right] \stackrel{\text{Jensen's inequality}}{\le} \frac{1}{\tau^4}\mathbb{E}_{x\sim \overline{X}_\tau}\left[\mathbb{E}\left[\|Z\|_2^8|\overline{X}_\tau= x\right]\right]\notag\\
			& = \frac{1}{\tau^4}\mathbb{E}[\|Z\|_2^8] \asymp \frac{d^4}{\tau^4}.\label{ineq:technical_2}
		\end{align}

\section{Proof of lower bound (Theorem \ref{thm:lower_bound})}
\label{sec:proof-thm-lower-bound}

We know from \eqref{eq:forward} that $X_t \sim N(0, \lambda_tI_d)$, where
\begin{align}\label{eq:lambda_t_recursion}
	\lambda_t = \alpha_t\lambda_{t-1} + 1 - \alpha_t,~\quad~\forall 1 \leq t \leq T,~\quad~\lambda_0 = \lambda.
\end{align}
Solving \eqref{eq:lambda_t_recursion} gives us
\begin{align}\label{eq:lambda_t}
	\lambda_t = (1 - \overline{\alpha}_t) + \overline{\alpha}_t\lambda,~\quad~\forall 1 \leq t \leq T.
\end{align}
By virtue of \eqref{eq:DDPM} and the fact $s_t^\star(x) = \nabla \log p_{X_t}(x) = -\frac{x}{\lambda_t}$, it is straightforward to check that
\begin{align}\label{eq:Y_t_form}
	Y_t \sim N(0, \widehat{\lambda}_tI_d),
\end{align}
where the variances $\{\widehat{\lambda}_t\}_{t=1}^T$ satisfy $\widehat{\lambda}_{T} = 1$ and 
\begin{align*}
	\widehat{\lambda}_{t-1} 
	&= \frac{\widehat{\lambda}_t}{\alpha_t}\left(1-\frac{\eta_t}{\lambda_t}\right)^2 + \frac{\sigma_t^2}{\alpha_t},~\quad~\forall 2 \leq t \leq T.
\end{align*}
Unrolling the recursion gives
\begin{align}\label{eq:widehat_lambda_t}
	\widehat{\lambda}_{1}=\sum_{t=2}^T \prod_{i=2}^{t-1}\frac{1}{\alpha_i}\left(1-\frac{\eta_i}{\lambda_i}\right)^2\frac{\sigma_t^2}{\alpha_t}.
\end{align}
Then we can write the close form of the KL divergence between $X_1$ and $Y_1$ admits the following closed-form expression:
\begin{align*}
	\mathsf{KL}(p_{X_1}||p_{Y_1}) = \frac{d}{2}\left(\log\frac{\widehat{\lambda}_{1}}{{\lambda}_{1}} + \frac{\lambda_{1}}{\widehat{\lambda}_{1}} - 1\right).
\end{align*} 
For ${\lambda}_{1}/\widehat{\lambda}_{1}\in(0,2)$, using the inequality $-\log x\ge (1-x) + \frac14(x-1)^2$ for $0 < x < 2$, we have 
$$
\mathsf{KL}(p_{X_1}||p_{Y_1}) \ge \frac{d}{8} \left(1-\frac{{\lambda}_{1}}{\widehat{\lambda}_{1}}\right)^2.
$$
For ${\lambda}_{1}/\widehat{\lambda}_{1}\ge 2$, the inequality $\log x\le (x-1)\log2$ for $x \geq 2$ yields
$$
\mathsf{KL}(p_{X_1}||p_{Y_1}) \ge \frac{d}{2}\left(\left(1-\frac{\lambda_{1}}{\widehat{\lambda}_{1}}\right)\log 2 + \frac{\lambda_{1}}{\widehat{\lambda}_{1}} - 1\right) = \frac{d(1-\log 2)}{2}\left(\frac{\lambda_{1}}{\widehat{\lambda}_{1}} - 1\right).
$$
Therefore, it suffices to prove that $\left|\frac{{\lambda}_{1}}{\widehat{\lambda}_{1}} - 1\right| \ge \frac{C}{T}$ for some small constant $C$, which can be guaranteed by
$$
\left|\frac{\widehat{\lambda}_{1}}{\lambda_{1}} - 1\right| \ge \frac{2C}{T}
$$
as long as $T\gg C$.

We first focus on the case $\eta_t = 1-\alpha_t$ and $\sigma_t^2 = 1-\alpha_t$.
Note that when $\eta_t = 1-\alpha_t$, we have
$$
1-\frac{\eta_t}{\lambda_t} = 1-\frac{1-\alpha_t}{\lambda_t} \stackrel{\eqref{eq:lambda_t_recursion}}{=} \frac{\alpha_t\lambda_{t-1}}{\lambda_t}.
$$
Putting the previous equation and \eqref{eq:widehat_lambda_t} together, one has
\begin{align*}
\widehat{\lambda}_1
&=\sum_{t=2}^T \prod_{i=2}^{t-1}\frac{1}{\alpha_i}\left(\frac{\alpha_i\lambda_{i-1}}{\lambda_i}\right)^2\frac{\sigma_t^2}{\alpha_t}
=\sum_{t=2}^T \frac{\lambda_1^2}{\alpha_1}\frac{\overline{\alpha}_{t-1}\sigma_t^2}{\alpha_t\lambda_{t-1}^2}\notag\\
&=\frac{\overline{\alpha}_{T}\lambda_1^2}{\alpha_1\lambda_T^2}\widehat{\lambda}_T 
+ \frac{\lambda_1^2}{\alpha_1}\sum_{t=1}^{T-1}\frac{\overline{\alpha}_t(1-\alpha_{t+1})}{\lambda_t^2\alpha_{t+1}} 
= \frac{\overline{\alpha}_{T}\lambda_1^2}{\alpha_1\lambda_T^2}\widehat{\lambda}_T 
+ \frac{\lambda_1^2}{\alpha_1}\sum_{t=1}^{T-1}\frac{\overline{\alpha}_t-\overline{\alpha}_{t+1}}{\lambda_t^2\alpha_{t+1}}.
\end{align*}
In addition, we can decompose $\lambda_1$ as 
\begin{align*}
\lambda_1 = \frac{\lambda_1^2}{\alpha_1}\cdot\frac{\overline{\alpha}_1}{\lambda_1} = \frac{\lambda_1^2\overline{\alpha}_{T}}{\alpha_1\lambda_T} 
+ \frac{\lambda_1^2}{\alpha_1}\sum_{t=1}^{T-1}\left(\frac{\overline{\alpha}_t}{\lambda_t} - \frac{\overline{\alpha}_{t+1}}{\lambda_{t+1}}\right)
\stackrel{\eqref{eq:lambda_t_recursion}}{=}\frac{\overline{\alpha}_{T}\lambda_1^2}{\alpha_1\lambda_T^2}\lambda_T
+ \frac{\lambda_1^2}{\alpha_1}\sum_{t=1}^{T-1}\frac{\overline{\alpha}_t-\overline{\alpha}_{t+1}}{\lambda_t\lambda_{t+1}}.
\end{align*}
The previous two equations together imply
\begin{align}\label{eq:expression_lambda_differencce}
\frac{\widehat{\lambda}_1}{\lambda_1} - 1 
&= \frac{\widehat{\lambda}_1 - \lambda_1}{\lambda_1}\notag\\
&= \frac{\overline{\alpha}_{T}\lambda_1}{\alpha_1\lambda_T^2}\widehat{\lambda}_T 
+ \frac{\lambda_1}{\alpha_1}\sum_{t=1}^{T-1}\frac{\overline{\alpha}_t-\overline{\alpha}_{t+1}}{\lambda_t^2\alpha_{t+1}} - \left(\frac{\overline{\alpha}_{T}\lambda_1}{\alpha_1\lambda_T^2}\lambda_T
+ \frac{\lambda_1}{\alpha_1}\sum_{t=1}^{T-1}\frac{\overline{\alpha}_t-\overline{\alpha}_{t+1}}{\lambda_t\lambda_{t+1}}\right)\notag\\
&= \frac{\overline{\alpha}_{T}\lambda_1}{\alpha_1\lambda_T^2}(\widehat{\lambda}_T - \lambda_T) 
+ \frac{\lambda_1}{\alpha_1}\sum_{t=1}^{T-1}\frac{\overline{\alpha}_t-\overline{\alpha}_{t+1}}{\lambda_t}\left(\frac{1}{\lambda_t\alpha_{t+1}}-\frac{1}{\lambda_{t+1}}\right)\notag\\
&\stackrel{\eqref{eq:lambda_t_recursion}}{=} \frac{\overline{\alpha}_{T}\lambda_1}{\alpha_1\lambda_T^2}(\widehat{\lambda}_T - \lambda_T) 
+ \frac{\lambda_1}{\alpha_1}\sum_{t=1}^{T-1}\frac{\overline{\alpha}_t-\overline{\alpha}_{t+1}}{\lambda_t}\frac{1-\alpha_{t+1}}{\lambda_t\lambda_{t+1}\alpha_{t+1}}\notag\\
&=\frac{\overline{\alpha}_{T}\lambda_1}{\alpha_1\lambda_T^2}(\widehat{\lambda}_T - \lambda_T) 
+ \frac{\lambda_1}{\alpha_1}\sum_{t=1}^{T-1}\frac{(\overline{\alpha}_t-\overline{\alpha}_{t+1})^2}{\lambda_t^2\lambda_{t+1}\overline{\alpha}_{t+1}}.
\end{align}
If $\lambda \asymp 1$, then we have $\lambda_t \asymp 1$ for all $t$. Applying Cauchy-Schwarz inequality and the fact $\overline{\alpha}_t \leq 1$ for all $t$ yields
\begin{align}\label{eq:lower_bound_1}
	\frac{\lambda_1}{\alpha_1}\sum_{t=1}^{T-1}\frac{(\overline{\alpha}_t-\overline{\alpha}_{t+1})^2}{\lambda_t^2\lambda_{t+1}\overline{\alpha}_{t+1}} \asymp \frac{\lambda_1}{\alpha_1}\sum_{t=1}^{T-1}\frac{(\overline{\alpha}_t-\overline{\alpha}_{t+1})^2}{\overline{\alpha}_{t+1}} \geq \frac{\lambda_1}{\alpha_1}\frac{\left(\sum_{t=1}^{T-1}(\overline{\alpha}_t-\overline{\alpha}_{t+1})\right)^2}{\sum_{t=1}^{T-1}\overline{\alpha}_{t+1}} \geq \frac{\lambda_1(\overline{\alpha}_1-\overline{\alpha}_T)^2}{T\alpha_1} = \frac{\lambda_1(\alpha_1-\overline{\alpha}_T)^2}{T\alpha_1}.
\end{align}
This, together with \eqref{eq:learning-rate}, tells us that
\begin{align}\label{eq:lower_bound_part_1}
	\frac{\lambda_1}{\alpha_1}\sum_{t=1}^{T-1}\frac{(\overline{\alpha}_t-\overline{\alpha}_{t+1})^2}{\lambda_t^2\lambda_{t+1}\overline{\alpha}_{t+1}} \gtrsim \frac{\lambda_1(1 - \frac{1}{T^{c_0}}-\frac{1}{T^{c_0}})^2}{T} \asymp \frac{1}{T}.
\end{align}
Moreover, in view of \eqref{eq:lambda_t} and the fact $\widehat{\lambda}_T = 1$, we have
\begin{align}\label{eq:lower_bound_part_2}
	\left|\frac{\overline{\alpha}_{T}\lambda_1}{\alpha_1\lambda_T^2}(\widehat{\lambda}_T - \lambda_T)\right| = \frac{\overline{\alpha}_{T}^2\lambda_1}{\alpha_1\lambda_T^2}(\lambda - 1) \asymp \frac{\overline{\alpha}_{T}^2}{\alpha_1} \stackrel{\eqref{eq:learning-rate}}{\gtrsim} \frac{1}{T^{2c_0}}.
\end{align}
Combining \eqref{eq:expression_lambda_differencce}, \eqref{eq:lower_bound_part_1} and \eqref{eq:lower_bound_part_2}, we have
\begin{align*}
	\left|\frac{\overline{\alpha}_{T}\lambda_1}{\alpha_1\lambda_T^2}(\widehat{\lambda}_T - \lambda_T)\right| \geq \frac{\lambda_1}{\alpha_1}\sum_{t=1}^{T-1}\frac{(\overline{\alpha}_t-\overline{\alpha}_{t+1})^2}{\lambda_t^2\lambda_{t+1}\overline{\alpha}_{t+1}} - \left|\frac{\overline{\alpha}_{T}\lambda_1}{\alpha_1\lambda_T^2}(\widehat{\lambda}_T - \lambda_T)\right| \gtrsim \frac{1}{T}.
\end{align*}
This completes the proof.

For $\sigma_t^2 = (1-\alpha_t)\alpha_t$ and $\frac{(\alpha_t-\overline{\alpha}_t)(1-{\alpha}_{t})}{1-\overline{\alpha}_t}$, the desired bound follows by similar arguments. We omit the details here for the sake of brevity.

\section{Proof of Examples}
\label{app:proof-examples}

\paragraph{Proof of Example \ref{example:gauss}.} 
In this case, we can verify that
$$
\tau \nabla \overline{s}^{\star}_\tau(x) = -\tau ((1-\tau)\Sigma + \tau I)^{-1},
$$
and thus $\|\tau \nabla \overline{s}^{\star}_\tau(x)\|_{\mathsf{op}}\le 1$.

\paragraph{Proof of Example \ref{example:GMM}.}

For ease of notation, we define the adjusted mean and variance by
$$
\widetilde{\sigma}_h^2\coloneqq (1-\tau)\sigma_h^2+\tau,\qquad \widetilde{\mu}_h\coloneqq \sqrt{1-\tau}\mu_h.
$$
Then the posterior probability of the $h$-th Gaussian component is given by
$$
w_h(x)\coloneqq \frac{\pi_h\phi(x~|~\widetilde{\mu}_h,\widetilde{\sigma}_h^2I_d)}{\sum_{h'=1}^H\pi_{h'}\phi(x~|~\widetilde{\mu}_{h'},\widetilde{\sigma}_{h'}^2I_d)},
$$
where $\phi(\cdot~|~\widetilde{\mu}_h,\widetilde{\sigma}_h^2I_d)$ denotes the density of a Gaussian distribution with mean vector $\widetilde{\mu}_h$ and covariance matrix $\widetilde{\sigma}_h^2I_d$. For any $h \in [H]$, we define $$z_h(x) := \frac{x - \widetilde{\mu}_h}{\widetilde{\sigma}_h}.$$ Then it is straightforward to show that
\begin{align}\label{eq:score_gaussian_mixture}
	\overline{s}^{\star}_\tau(x) = \nabla \log p_{\overline{X}_{\tau}}(x) = \nabla \log\left(\sum_{h=1}^{H}\pi_h\phi\left(x~|~\widetilde{\mu}_h,\widetilde{\sigma}_h^2I_d\right)\right) = -\sum_{h=1}^{H}w_h(x)\frac{x - \widetilde{\mu}_h}{\widetilde{\sigma}_h^2} = -\sum_{h=1}^{H}w_h(x)\frac{z_h(x)}{\widetilde{\sigma}_h},
\end{align}
which further implies
\begin{align}\label{eq:Hessian}
\nabla \overline{s}^{\star}_\tau(x) 
&= -\sum_{h=1}^H\frac{w_h(x)}{\widetilde{\sigma}_h^2}I_d + \sum_{h=1}^Hw_h(x)\frac{x - \widetilde{\mu}_h}{\widetilde{\sigma}_h^2}\left(\frac{x - \widetilde{\mu}_h}{\widetilde{\sigma}_h^2}\right)^\top - \left(\sum_{h=1}^Hw_h(x)\frac{x - \widetilde{\mu}_h}{\widetilde{\sigma}_h^2}\right)\left(\sum_{h=1}^Hw_h(x)\frac{x - \widetilde{\mu}_h}{\widetilde{\sigma}_h^2}\right)^\top\notag\\
&= -\sum_{h=1}^H\frac{w_h(x)}{\widetilde{\sigma}_h^2}I_d + \sum_{h=1}^Hw_h(x)\left(\frac{z_h(x)}{\widetilde{\sigma}_h}-\sum_{h'=1}^H\frac{w_{h'}(x)z_{h'}(x)}{\widetilde{\sigma}_{h'}}\right)\left(\frac{z_h(x)}{\widetilde{\sigma}_h}-\sum_{h'=1}^H\frac{w_{h'}(x)z_{h'}(x)}{\widetilde{\sigma}_{h'}}\right)^{\top}.
\end{align}
We make the observation that for any vectors $u_1, \cdots, u_H, h \in \mathbb{R}^d$ and $w_1, \cdots, w_H \geq 0$ satisfying $\sum_{h=1}^{H}w_h = 1$, 
\begin{align*}
	\sum_{h=1}^{H}w_h\left(u_h - u\right)\left(u_h - u\right)^\top &= \sum_{h=1}^{H}w_h\left(u_h - \sum_{i=1}^{H}w_iu_i + \sum_{i=1}^{H}w_iu_i - u\right)\left(u_h - \sum_{i=1}^{H}w_iu_i + \sum_{i=1}^{H}w_iu_i - u\right)^\top\\
	&= \sum_{h=1}^{H}w_h\left(u_h - \sum_{i=1}^{H}w_iu_i\right)\left(u_h - \sum_{i=1}^{H}w_iu_i\right)^\top + \left[\sum_{h=1}^{H}w_h\left(u_h - \sum_{i=1}^{H}w_iu_i\right)\right]\left(\sum_{i=1}^{H}w_iu_i - u\right)^\top\\
	&\quad + \left(\sum_{i=1}^{H}w_iu_i - u\right)\left[\sum_{h=1}^{H}w_h\left(u_h - \sum_{i=1}^{H}w_iu_i\right)^\top\right] + \sum_{h=1}^{H}w_h\left(\sum_{i=1}^{H}w_iu_i - u\right)\left(\sum_{i=1}^{H}w_iu_i - u\right)^\top\\
	&= \sum_{h=1}^{H}w_h\left(u_h - \sum_{i=1}^{H}w_iu_i\right)\left(u_h - \sum_{i=1}^{H}w_iu_i\right)^\top + \sum_{h=1}^{H}w_h\left(\sum_{i=1}^{H}w_iu_i - u\right)\left(\sum_{i=1}^{H}w_iu_i - u\right)^\top\\
	&\succeq \sum_{h=1}^{H}w_h\left(u_h - \sum_{i=1}^{H}w_iu_i\right)\left(u_h - \sum_{i=1}^{H}w_iu_i\right)^\top.
\end{align*}
Here, $A \succeq B$ means that $A - B$ is positive semi-definite.
By virtue of the previous inequality, we know that for any $k \in \{1, \dots, H\}$, 
\begin{align}\label{eq:Hessian-upper}
&\sum_{h=1}^Hw_h(x)\left(\frac{z_h(x)}{\widetilde{\sigma}_h}-\sum_{h'=1}^H\frac{w_{h'}(x)z_{h'}(x)}{\widetilde{\sigma}_{h'}}\right)\left(\frac{z_h(x)}{\widetilde{\sigma}_h}-\sum_{h'=1}^H\frac{w_{h'}(x)z_{h'}(x)}{\widetilde{\sigma}_{h'}}\right)^{\top}\notag\\
&\quad \preceq \sum_{h=1}^Hw_h(x)\left(\frac{z_h(x)}{\widetilde{\sigma}_h}-\frac{z_k(x)}{\widetilde{\sigma}_k}\right)\left(\frac{z_h(x)}{\widetilde{\sigma}_h}-\frac{z_k(x)}{\widetilde{\sigma}_k}\right)^{\top}\notag\\
&\quad = \widetilde{\sigma}_k^{-2}\sum_{h=1}^Hw_h(x)\left(\frac{z_h(x)}{\overline{\sigma}_{h,k}}-{z}_{k}(x)\right)\left(\frac{z_h(x)}{\overline{\sigma}_{h,k}}-{z}_k(x)\right)^{\top},
\end{align}
where
$\overline{\sigma}_{h,k} \coloneqq \widetilde{\sigma}_h/\widetilde{\sigma}_{k}$.
Recalling that $\overline{X}_\tau$ follows the GMM $\sum_{k=1}^H \pi_k\mathcal{N}(\widetilde{\mu}_k, \widetilde{\sigma}_k^2I_d)$,
we have
\begin{align*}
&\mathbb{P}_{X \sim \overline{X}_\tau}\left(\|\tau \nabla \overline{s}_\tau^{\star}(X)\|_{\mathsf{op}}\ge C\log H\log d\right)\notag\\
&\quad= \sum_{k=1}^H\pi_k\mathbb{P}_{X\sim \mathcal{N}(\widetilde{\mu}_k,\widetilde{\sigma}_k^2I_d)}\left(\|\tau \nabla \overline{s}_\tau^{\star}(X)\|_{\mathsf{op}}\ge C\log H\log d\right)\notag\\
&\quad= \sum_{k: \pi_k\ge 1/(2d^4H)}\pi_k\mathbb{P}_{X \sim \mathcal{N}(\widetilde{\mu}_k,\widetilde{\sigma}_k^2I_d)}\left(\|\tau \nabla \overline{s}_\tau^{\star}(X)\|_{\mathsf{op}}\ge C\log H\log d\right)\notag\\&\qquad + \sum_{k: \pi_k > 1/(2d^4H)}\pi_k\mathbb{P}_{X\sim \mathcal{N}(\widetilde{\mu}_k,\widetilde{\sigma}_k^2I_d)}\left(\|\tau \nabla \overline{s}_\tau^{\star}(X)\|_2\ge C\log H\log d\right)\notag\\
&\quad\le \sum_{\pi_k\ge 1/(2d^4H)}\pi_k\mathbb{P}_{X \sim \mathcal{N}(\widetilde{\mu}_k,\widetilde{\sigma}_k^2I_d)}\left(\|\tau \nabla \overline{s}_\tau^{\star}(X)\|_{\mathsf{op}}\ge C\log H\log d\right) + \frac{1}{2d^4}.
\end{align*}
Noting that $\widetilde{\sigma}_h^2 \ge \tau$ for any $h$, we obtain
$$
\tau \sum_{h=1}^H \frac{w_h(x)}{\widetilde{\sigma}_h^2}I_d \preceq \sum_{h=1}^Hw_h(x) I_d = I_d.
$$
By virtue of \eqref{eq:Hessian}, \eqref{eq:Hessian-upper} and the previous two inequalities,
we further have
\begin{align*}
&\quad\mathbb{P}_{X\sim \overline{X}_\tau}\left(\|\tau \nabla \overline{s}_\tau^{\star}(X)\|_{\mathsf{op}}\ge C\log H\log d\right)\notag\\
&\le \sum_{\pi_k\ge 1/(2d^4H)}\pi_k\mathbb{P}_{X \sim \mathcal{N}(\widetilde{\mu}_k,\widetilde{\sigma}_k^2I_d)}\left(\left\|\sum_{h=1}^Hw_h(X)\left(\frac{z_h(X)}{\overline{\sigma}_{h,k}}-{z}_{k}(X)\right)\left(\frac{z_h(X)}{\overline{\sigma}_{h,k}}-{z}_k(X)\right)^{\top}\right\|_{\mathsf{op}}\ge \frac{C}{2}\log H\log d\right) + \frac{1}{2d^4}\notag\\
&\le \sum_{\pi_k\ge 1/(2d^4H)}\pi_k\mathbb{P}_{X\sim \mathcal{N}(\widetilde{\mu}_k,\widetilde{\sigma}_k^2I_d)}\left(\sum_{h=1}^Hw_h(X)\left\|\frac{z_h(X)}{\overline{\sigma}_{h,k}}-{z}_{k}(X)\right\|_2^2 \ge \frac{C}{2}\log H\log d\right) + \frac{1}{2d^4},
\end{align*}
where the last inequality uses the fact
\begin{align*}
\left\|\sum_{h=1}^Hw_h(x)\left(\frac{z_h(x)}{\overline{\sigma}_{h,k}}-{z}_{k}(x)\right)\left(\frac{z_h(x)}{\overline{\sigma}_{h,k}}-{z}_k(x)\right)^{\top}\right\|_{\mathsf{op}}
\le \sum_{h=1}^Hw_h(x)\left\|\frac{z_h(x)}{\overline{\sigma}_{h,k}}-{z}_{k}(x)\right\|_2^2.
\end{align*}
To complete the proof, it suffices to show that, for any component $k$ with prior probability $\pi_k \ge 1/(2d^4H)$,
if $X \sim \mathcal{N}(\widetilde{\mu}_k,\widetilde{\sigma}_k^2I)$ (so that $z_k(X)\sim \mathcal{N}(0,I_d)$), then
the following holds with probability at least $1-1/(2d^4)$:
\begin{align}\label{eq:proof-example-GMM-reduce}
\sum_{h=1}^Hw_h(X)\left\|\frac{z_h(X)}{\overline{\sigma}_{h,k}}-{z}_{k}(X)\right\|_2^2 \lesssim \log(dH).
\end{align}
If $d\leq C_0\log(dH)$ for some sufficiently large constant $C-0 > 0$, since $\log(d)\log(H)\geq d \log d$, Example \ref{example:GMM} follows directly from \cite[Lemma 1]{li2024d}.
In the remaining analysis, we focus on the case $d\geq C_0\log(dH)$.
For convenience, we let $$\gamma_{h,k}\coloneqq \overline{\sigma}_{h,k}^{-2}-1\in(-1,\infty).$$ We claim that, for any $k$ satisfying $\pi_k \ge 1/(2d^4H)$ and for each $h \in\{1, \cdots, H\}$, 
\begin{align}\label{eq:example_1_key}
	\mathbb{P}_{X \sim N(\widetilde{\mu}_k, \widetilde{\sigma}_k^2I_d)}\left(\left\|\frac{z_h(X)}{\overline{\sigma}_{h,k}}-z_k(X)\right\|_2^2 \lesssim \max\left\{\log(dH), \frac{w_k(X)}{w_h(X)}\right\}\right) \geq 1 - \frac{1}{2d^4H}.
\end{align}
If \eqref{eq:example_1_key} holds, then it together with the union bound shows that with probability exceeding $1 - 1/(2d^{4})$,
\begin{align*}
	\sum_{h=1}^Hw_h(X)\left\|\frac{z_h(X)}{\overline{\sigma}_{h,k}}-{z}_{k}(X)\right\|_2^2 &\lesssim \sum_{h=1}^Hw_h(X)\left(\log(dH) + \frac{w_k(X)}{w_h(X)}\right)\\
	&= \log(dH)\sum_{h=1}^Hw_h(X) + \sum_{h=1}^Hw_k(X) = \log(dH) +1,
\end{align*}
which confirms \eqref{eq:proof-example-GMM-reduce} and thus finishes the proof. Therefore, we only need to prove \eqref{eq:example_1_key}. 

To validate \eqref{eq:example_1_key}, we proceed by considering the following two scenarios.
\paragraph{Case 1: both $\|\widetilde{\mu}_k-\widetilde{\mu}_h\|/\widetilde{\sigma}_k$ and $|\gamma_{h,k}|$ are small.} We first consider the case that $\|\widetilde{\mu}_k-\widetilde{\mu}_h\|/\widetilde{\sigma}_k \le C_1\sqrt{\log (dH)}$ and $|\gamma_{h,k}|\le C_2\sqrt{\frac{\log (dH)}{d}}$ for some sufficiently large constants $C_1, C_2 > 0$. For any $x$ satisfying $\|z_k(x)\|_2^2 \lesssim d$, we have
\begin{align}
	\left\|\frac{z_h(x)}{\overline{\sigma}_{h,k}}-z_k(x)\right\|_2^2
	&=\left\|\frac{1}{\overline{\sigma}_{h,k}}\left(\frac{z_k(x)}{\overline{\sigma}_{h,k}}+\frac{1}{\overline{\sigma}_{h,k}}\frac{\widetilde{\mu}_k-\widetilde{\mu}_h}{\widetilde{\sigma}_k}\right)-z_k(x)\right\|_2^2\notag\\
	&=\left\|\gamma_{h,k} z_k(x) + (\gamma_{h,k}+1)\frac{\widetilde{\mu}_k-\widetilde{\mu}_h}{\widetilde{\sigma}_k}\right\|_2^2\notag\\
	&\lesssim \gamma_{h,k}^2\left\|z_k(x)\right\|_2^2 + (\gamma_{h,k}^2+1)\left\|\frac{\widetilde{\mu}_k-\widetilde{\mu}_h}{\widetilde{\sigma}_k}\right\|_2^2\label{eq:example_2_2}\\ 
	&\lesssim \frac{\log(dH)}{d}\cdot d + \left(\frac{\log(dH)}{d}+1\right)\log(dH)\notag\\&\lesssim \log(dH)\label{eq:example_2_1},
\end{align}
provided that $d\geq \log(H)$. Moreover, by virtue of \cite[Lemma 1]{laurent2000adaptive}, we have
\begin{align*}
	\mathbb{P}_{X \sim N(\widetilde{\mu}_k, \widetilde{\sigma}_k^2I_d)}\left(\|z_k(X)\|_2^2 \leq 13d\right) = \mathbb{P}_{z_k(X) \sim N(0, I_d)}\left(\|z_k(X)\|_2^2 \leq 13d\right) \geq 1 - e^{-4d} \geq 1 - \frac{1}{2Hd^4},
\end{align*}
provided that $d \geq \log(H)$. Putting \eqref{eq:example_2_1} and the previous inequality together, we know that \eqref{eq:example_1_key} holds.
\paragraph{Case 2: $\|\widetilde{\mu}_k-\widetilde{\mu}_h\|/\widetilde{\sigma}_k$ or $|\gamma_{h,k}|$ is large.}Now, we turn to the scenario that $\|\widetilde{\mu}_k-\widetilde{\mu}_h\|/\widetilde{\sigma}_k \ge C_1\sqrt{\log (dH)}$ or $|\gamma_{h,k}|\ge C_2\sqrt{\frac{\log (dH)}{d}}$.
It suffices to show that there exists a constant $C' > 0$ such that, with probability at least $1-1/(4d^4H)$,
\begin{align}\label{eq:proof-example-gmm-reduce-2}
	\log\left(\frac{w_h(X)}{C' w_k(X)}\left\|\frac{z_h(X)}{\overline{\sigma}_{h,k}}-z_k(X)\right\|_2^2\right)\le 0.
\end{align}
To establish \eqref{eq:proof-example-gmm-reduce-2}, we first compute
\begin{align*}
	\frac{w_h(x)}{w_k(x)} &= \frac{\pi_h\phi(x~|~\widetilde{\mu}_h,\widetilde{\sigma}_h^2I_d)}{\pi_k\phi(x~|~\widetilde{\mu}_k,\widetilde{\sigma}_k^2I_d)}\\
	&= \exp\left(-\frac{\|z_h(x)\|_2^2}{2} + \frac{\|z_k(x)\|_2^2}{2} + \frac{d}{2}\log\frac{\widetilde{\sigma}_k^2}{\widetilde{\sigma}_h^2} + \log\frac{\pi_h}{\pi_k}\right)\notag\\
	&=\exp\left(
	-\frac{\gamma_{h,k}\|z_k(x)\|_2^2}{2}
	- \frac{(\gamma_{h,k}+1)}{2\widetilde{\sigma}_k^2}\|\widetilde{\mu}_k-\widetilde{\mu}_h\|_2^2 - \frac{z_k(x)^{\top}(\widetilde{\mu}_k-\widetilde{\mu}_h)}{\widetilde{\sigma}_k\overline{\sigma}_{h,k}^2} + \frac{d}{2}\log(\gamma_{h,k}+1) + \log\frac{\pi_h}{\pi_k}\right).
\end{align*}
Since $\pi_k \ge 1/(2d^4H)$ and $H\ge 2$, we have
\begin{align*}
	\log\frac{\pi_h}{\pi_k}\le 4\log(dH).
\end{align*}
In view of \cite[Lemma 1]{laurent2000adaptive}, the standard Gaussian concentration inequality and the fact that $z_k(X) \sim N(0, I_d)$, we know that, with probability exceeding $1-1/(4d^4H)$,
\begin{align}\label{eq:proof-examples-GMM-probset}
	\left\|z_k(X)\right\|_2^2 - d\lesssim \sqrt{d\log(dH)},\qquad \mathrm{and}\qquad |{z}_k(X)^{\top}(\widetilde{\mu}_k-\widetilde{\mu}_h)|\lesssim \|\widetilde{\mu}_k-\widetilde{\mu}_h\|_2\sqrt{\log(dH)}.
\end{align}
We define the event $\mathcal{E} := \{\eqref{eq:proof-examples-GMM-probset}~\text{holds}\}$. On $\mathcal{E}$, one has
\begin{align}\label{eq:example_2_3}
	&\log\left(\frac{w_h(X)}{C' w_k(X)}\left\|\frac{z_h(X)}{\overline{\sigma}_{h,k}}-z_k(X)\right\|_2^2\right)\notag\\
	&\quad\stackrel{\eqref{eq:example_2_2}~\text{and}~\eqref{eq:proof-examples-GMM-probset}}{\leq} -\frac{\gamma_{h,k}d-C_3|\gamma_{h,k}|\sqrt{d\log(dH)}}{2}
	- \frac{(\gamma_{h,k}+1)}{2\widetilde{\sigma}_k^2}\|\widetilde{\mu}_k-\widetilde{\mu}_h\|_2^2 \notag\\
	&\qquad + \frac{C_4(\gamma_{h,k}+1)\|\widetilde{\mu}_k-\widetilde{\mu}_h\|_2\sqrt{\log(dH)}}{\widetilde{\sigma}_k}\notag\\
	&\qquad + \frac{d}{2}\log(\gamma_{h,k}+1) + 4\log(dH) + \log\left(\gamma_{h,k}^2 d\right) + \log\left((\gamma_{h,k}+1)^2\frac{\|\widetilde{\mu}_k-\widetilde{\mu}_h\|_2^2}{C'\widetilde{\sigma}_k^2}\right)\notag\\
	&\quad\overset{\text{(i)}}{\le}- \frac{\gamma_{h,k}d-C_3'|\gamma_{h,k}|\sqrt{d\log(dH)}}{2}
	- \frac{(\gamma_{h,k}+1)}{2\widetilde{\sigma}_k^2}\|\widetilde{\mu}_k-\widetilde{\mu}_h\|_2^2\notag\\
	&\qquad + \frac{C_4'(\gamma_{h,k}+1)\|\widetilde{\mu}_k-\widetilde{\mu}_h\|_2\sqrt{\log(dH)}}{\widetilde{\sigma}_k} + \frac{d}{2}\log(\gamma_{h,k}+1) + 4\log(dH),
\end{align}
where $C_3$ and $C_4$ are some universal constants,
$C_3' = C_3 + 4$, $C_4' = C+2/\sqrt{C'}$,
and (i) comes from $\log(\gamma^2 d)=2\log(|\gamma| \sqrt{d})\le 2|\gamma| \sqrt{d}$
and 
$$
\log\left((\gamma_{h,k}+1)^2\frac{\|\widetilde{\mu}_k-\widetilde{\mu}_h\|_2^2}{C'\widetilde{\sigma}_k^2}\right)
\le \frac{2}{\sqrt{C'}}(\gamma_{h,k}+1)^2\frac{\|\widetilde{\mu}_k-\widetilde{\mu}_h\|_2}{\widetilde{\sigma}_k}.
$$
To prove \eqref{eq:proof-example-gmm-reduce-2}, we consider three cases.
\begin{itemize}
	\item[Case 2.1.] 
	If $\|\widetilde{\mu}_k-\widetilde{\mu}_h\|_2/\widetilde{\sigma}_k \ge C_1 \sqrt{\log (dH)}$ and $|\gamma_{h,k}|\ge C_2 \sqrt{\log(dH)/d}$, 
	then we have
	\begin{align}\label{eq:proof-example-GMM-muterm-1}
		- \frac{(\gamma_{h,k}+1)}{2\widetilde{\sigma}_k^2}\|\widetilde{\mu}_k-\widetilde{\mu}_h\|_2^2 + \frac{C_4'(\gamma_{h,k}+1)\|\widetilde{\mu}_k-\widetilde{\mu}_h\|_2\sqrt{\log(dH)}}{\widetilde{\sigma}_k} \le - \frac{(\gamma_{h,k}+1)}{4\widetilde{\sigma}_k^2}\|\widetilde{\mu}_k-\widetilde{\mu}_h\|_2^2,
	\end{align}
	In addition, by virtue of the fact that $\log(x+1)-x\le -\frac14|x|\min\{|x|,1\}$ for any $x>-1$, we have
	\begin{align}\label{eq:proof-example-GMM-sigmaterm-1}
		&-\frac{\gamma_{h,k}d-C_3'|\gamma_{h,k}|\sqrt{d\log(dH)}}{2}
		+ \frac{d}{2}\log(\gamma_{h,k}+1)\notag\\
		&\quad\le \frac{C_3'|\gamma_{h,k}|\sqrt{d\log(dH)}}{2}-\frac{d|\gamma_{h,k}|}{8}\min\{|\gamma_{h,k}|,1\}\notag\\
		&\quad\le \frac{d|\gamma_{h,k}|}{16}\sqrt{\frac{64C_3'^{2}\log(dH)}{d}}-\frac{d|\gamma_{h,k}|}{8}\min\{|\gamma_{h,k}|,1\}\notag\\
		&\quad\le \frac{d|\gamma_{h,k}|}{16}\min\left\{C_2\sqrt{\frac{\log(dH)}{d}},1\right\}-\frac{d|\gamma_{h,k}|}{8}\min\{|\gamma_{h,k}|,1\}\notag\\
		&\quad\le -\frac{d}{16}|\gamma_{h,k}|\min\{|\gamma_{h,k}|,1\},
	\end{align}
	provided that $C_2 \geq 8C_3'$ and $C_0 \geq 64C_3'^{2}$.
	Combining \eqref{eq:example_2_3}, \eqref{eq:proof-example-GMM-muterm-1} and \eqref{eq:proof-example-GMM-sigmaterm-1}, we obtain that on the event $\mathcal{E}$,
	\begin{align*}
		\log\left(\frac{w_h(X)}{C' w_k(X)}\left\|\frac{z_h(X)}{\overline{\sigma}_{h,k}}-z_k(X)\right\|_2^2\right)
		&\le -\frac{d}{16}|\gamma_{h,k}|\min\{|\gamma_{h,k}|,1\} - \frac{(\gamma_{h,k}+1)}{4\widetilde{\sigma}_k^2}\|\widetilde{\mu}_k-\widetilde{\mu}_h\|_2^2 + 4\log(dH)\\
		&\le -\frac{d}{16}C_2 \sqrt{\log(dH)/d}\min\left\{C_2 \sqrt{\log(dH)/d}, 1\right\} + 4\log(dH)\\
		&= \min\left\{-\frac{C_2^2}{16}\log(dH), -\frac{C_2}{16}\sqrt{d\log(dH)}\right\} + 4\log(dH)\le 0.
	\end{align*}
	\item[Case 2.2.]
	If $\|\widetilde{\mu}_k-\widetilde{\mu}_h\|/\widetilde{\sigma}_k \ge C_1 \sqrt{\log (dH)}$ and $|\gamma_{h,k}|\le C_2\sqrt{\log(dH)/d}$, 
	then \eqref{eq:example_2_3} tells us that on the event $\mathcal{E}$,
	\begin{align*}
		&\log\left(\frac{w_h(x)}{C' w_k(x)}\left\|\frac{z_h(x)}{\overline{\sigma}_{h,k}}-z_k(x)\right\|_2^2\right)\notag\\
		&\quad\le \frac{C_3'|\gamma_{h,k}|\sqrt{d\log(dH)}}{2} - \frac{(\gamma_{h,k}+1)}{4\widetilde{\sigma}_k^2}\|\mu_k-\mu_h\|_2^2 +4\log(dH) - \frac{d}{2}\left(\gamma_{h,k} - \log(\gamma_{h,k} + 1)\right)\notag\\
		&\quad \le \frac{C_3'|\gamma_{h,k}|\sqrt{d\log(dH)}}{2} - \frac{(\gamma_{h,k}+1)}{4\widetilde{\sigma}_k^2}\|\mu_k-\mu_h\|_2^2 +4\log(dH)\notag\\
		&\quad\le (C_2C_3'/2+4)\log(dH) - \frac{\|\mu_k-\mu_h\|_2^2}{4\widetilde{\sigma}_k^2} \le 0,
	\end{align*}
	provided that $C_1^2\ge 2C_2C_3'+16$.
	\item[Case 2.3.] If $|\gamma_{h,k}|\ge C_2\sqrt{\log(dH)/d}$ and $\|\widetilde{\mu}_k-\widetilde{\mu}_h\|/\widetilde{\sigma}_k \le C_1\sqrt{\log (dH)}$, 
	then \eqref{eq:example_2_3} together with \eqref{eq:proof-example-GMM-sigmaterm-1} shows that
	\begin{align*}
		\log\left(\frac{w_h(x)}{C' w_k(x)}\left\|\frac{z_h(x)}{\overline{\sigma}_{h,k}}-z_k(x)\right\|_2^2\right)
		&\le \frac{C_4'(\gamma_{h,k}+1)\|\widetilde{\mu}_k-\widetilde{\mu}_h\|_2\sqrt{\log(dH)}}{\widetilde{\sigma}_k} -\frac{d}{16}|\gamma_{h,k}|\min\{|\gamma_{h,k}|,1\} + 4\log(dH)\notag\\
		&\le (C_1C_4'\gamma_{h,k}+C_1C_4'+4)\log(dH)-\frac{d}{16}|\gamma_{h,k}|\min\{|\gamma_{h,k}|,1\}\le 0,
	\end{align*}
	provided that $C_2 \ge \max\{32(C_1C_4'+4),\sqrt{32(C_1C_4'+4)}\}$ and $C_0 \ge \max\{16C_1C_4',(32C_1C_4'/C_2)^2\}$.
\end{itemize}
Putting the results for all cases together, we conclude that \eqref{eq:example_1_key} holds, which completes the proof.

\paragraph{Proof of Example \ref{example:independent}.}
In this case, $\nabla \overline{s}^{\star}_\tau(x)$ is a diagonal matrix with the $i$-th diagonal entry given by
\begin{align*}
[\nabla \overline{s}^{\star}_\tau(x)]_{i,i} = -\frac{1}{\tau} + \frac{1}{\tau^2}\mathbb{E}[\|x_i-\sqrt{1-\tau}x_{0,i}\|_2^2|\overline{X}_{\tau,i} = x_i] - \frac{1}{\tau^2}\|\mathbb{E}[x_i-\sqrt{1-\tau}x_{0,i}|\overline{X}_{\tau,i} = x_i]\|_2^2,
\end{align*}
where $x_i$, $\overline{X}_{\tau,i}$, and $x_{0,i}$ denote the $i$-th entry of $x$, $\overline{X}_\tau$, and $x_0$, respectively.
Thus we have
\begin{align}\label{eq:example_inequality_1}
	\|\nabla \overline{s}^{\star}_\tau(x)\|_{\mathsf{op}} \le \max_i|[\nabla \overline{s}^{\star}_\tau(x)]_{i,i}| \le \frac{1}{\tau} + \frac{1}{\tau^2}\max_i \mathbb{E}[\|x_i-\sqrt{1-\tau}x_{0,i}\|_2^2|\overline{X}_{\tau,i} = x_i].
\end{align}

Repeating similar arguments as in \cite[Eqn.~(A.6)]{li2024d},
we know that for any $\delta\in(0,1)$ and $x_i$ satisfying $\log p_{\overline{X}_{\tau,i}}(x_i) \ge -\theta \log \frac{1}{\delta}$ with some large constant $\theta \geq 1$, 
\begin{align*}
\frac{1}{\tau^2}\mathbb{E}[\|x_i-\sqrt{1-\tau}x_{0,i}\|_2^2|\overline{X}_{\tau,i} = x_i] \lesssim \frac{\log \frac{1}{\delta}}{\tau}.
\end{align*}
Taking $\delta = d^{-1}$, we have
\begin{align}\label{eq:example_inequality_2}
\frac{1}{\tau^2}\mathbb{E}[\|x_i-\sqrt{1-\tau}x_{0,i}\|_2^2|\overline{X}_{\tau,i} = x_i]\lesssim \frac{\log d}{\tau},
\end{align}
provided that $\log p_{\overline{X}_{\tau,i}}(x_i) \ge -\theta \log d$.
Combining \eqref{eq:example_inequality_1} and \eqref{eq:example_inequality_2}, we know that if $\log p_{\overline{X}_{\tau,i}}(x_i) \ge -\theta \log d$ for all $i \in d$
\begin{align*}
\tau \|\nabla \overline{s}^{\star}_\tau(x)\|_{\mathsf{op}}\lesssim \log d.
\end{align*}
Repeating similar arguments as in \cite[proof of Eqn.~(A.28b)]{li2024d}, we have
\begin{align*}
\mathbb{P}(\log p_{\overline{X}_{\tau,i}}(\overline{X}_{\tau,i}) \ge -\theta \log d) \le \frac{1}{d^5},
\end{align*}
which, together with the union bound, completes the proof.

\bibliographystyle{apalike}
\bibliography{refs}

\end{document}